\pdfoutput=1
\documentclass{article}
\usepackage{arxiv}
\usepackage[utf8]{inputenc}
\usepackage[T1]{fontenc}
\usepackage{amsmath}
\usepackage{amsfonts}
\usepackage{graphicx}
\usepackage{booktabs}
\usepackage{float}
\usepackage{array}
\usepackage{enumitem}
\usepackage{xcolor}
\usepackage{microtype}
\usepackage[numbers,sort]{natbib}
\usepackage{url}
\usepackage{hyperref}
\usepackage{authblk}

\hypersetup{colorlinks=true,linkcolor=black,citecolor=black,urlcolor=blue!50!black}
\newcommand{\dev}[1]{\raisebox{-0.650em}{\includegraphics[height=1.6em]{#1}}}

\newcolumntype{L}[1]{>{\raggedright\arraybackslash}p{#1}}
\newcolumntype{R}[1]{>{\raggedleft\arraybackslash}p{#1}}

\title{Adapting Small Language Models to Romanized Nepali: Tokenizer Fertility, Reference Likelihood and Generation Metrics}
\author[1,*]{Pujan Pant}
\author[1]{Sagar Newpane}
\author[2]{Hari Sitaula}
\author[3,*]{Manish Sakhakarmy}
\affil[1]{Herald College Kathmandu, University of Wolverhampton, Naxal, Kathmandu 44600, Nepal}
\affil[2]{Department of Mathematical Sciences, Montana Technological University, Butte, MT 59701, USA}
\affil[3]{ASI Industrial, Billings, MT 59101, USA}
\affil[*]{Correspondence: \texttt{pantpujan017@gmail.com} (P.P.); \texttt{sakhakarmym@gmail.com} (M.S.)}
\date{}
\renewcommand{\shorttitle}{Adapting Small Language Models to Romanized Nepali}
\hypersetup{pdftitle={Adapting Small Language Models to Romanized Nepali: Tokenizer Fertility, Reference Likelihood and Generation Metrics},pdfauthor={Pujan Pant, Sagar Newpane, Hari Sitaula, Manish Sakhakarmy},pdfsubject={cs.CL},pdfkeywords={Romanized Nepali, transliteration, small language models, QLoRA, tokenizer fertility, bits per byte, evaluation}}
\begin{document}
\maketitle

\begin{abstract}
Romanized Nepali, Nepali written in Latin script, is common in digital communication but has no standard spelling and few resources, and it is unclear how reliably small open language models can be adapted to it. We adapt three 2–3B base models from different families, Llama-3.2-3B, Qwen2.5-3B and Gemma-2-2B, with QLoRA on 9,000 instruction examples produced by deterministic transliteration of Nepali instruction data. Each model is trained with one shared recipe and three seeds and evaluated on 1,000 held-out examples with reference-overlap metrics, byte-normalized reference likelihood (bits per byte, BPB) and an output audit, against zero-shot baselines under two prompts. Fine-tuning raises chrF++ from 10–14 to 29.0–30.6 and lowers BPB from 2.23–2.41 to 0.78–0.84; most zero-shot outputs (54–92\%) are English-like. Likelihood and overlap disagree on how far apart the models are. Llama-3.2-3B has the lowest BPB in every seed, 0.042 below Qwen2.5-3B (95\% CI 0.039–0.046), yet the two cannot be separated on chrF++ (30.51 and 30.61) or BLEU, even with paired item-level tests. Gemma-2-2B, whose tokenizer splits Romanized Nepali words least (2.72 tokens per word), scores worst on every measure. Which model gains most over its zero-shot baseline depends on the zero-shot prompt. In a supplementary comparison with an identical tokenizer, Llama-3.1-8B lowers BPB by 0.053 (0.050–0.056) relative to Llama-3.2-3B, while chrF++ and BLEU rise by only 0.68 and 0.58 points. Lower fertility and higher reference likelihood therefore did not identify the model whose outputs were closest to the references.
\end{abstract}


\section{Introduction}\label{sec:intro}

Nepali is written in Devanagari, yet much of its everyday digital use is in Latin script. More than half of the comments in a Nepali social-media corpus were transliterated or mixed \citep{niraula2021}, and users of a Nepali sexual and reproductive health chatbot wrote 61.0\% of their queries in Romanized Nepali, against 23.2\% in Devanagari and 15.8\% in English \citep[Fig. 4A]{sharma2026}. Romanized Nepali has no standard orthography: the same word can be spelled in several ways, and English words are often mixed in \citep{chaudhary2025}. Language models also handle romanized input less well than the native script. In a real-world triage setting, one strong model's F1 on Romanized Nepali messages was 9.5 points below its F1 on Devanagari messages \citep{khullar2025}.

There are two broad responses. One converts the input: spelling normalization improved sentiment classification of Romanized Nepali reviews \citep{chaudhary2025}, and transliterating comments back to Devanagari helped multilingual classifiers \citep{patel2026}. The other adapts a model to read and write Romanized Nepali directly, which suits generation, where the output itself must be in the script that users type. Small open models of 2–3B parameters make this route practical, because parameter-efficient methods let them be adapted on a single 16~GB GPU \citep{dettmers2023}, as in this study.

Choosing and evaluating such models is not straightforward. Candidate models differ at once in tokenizer, architecture, size and pretraining data. Tokenizer fertility, the number of subword tokens per word, is an appealing selection criterion, because text that is rare in a tokenizer's training data is split into many pieces \citep{rust2021,petrov2023}, but its relation to downstream performance is inconsistent across studies \citep{ali2024,goldman2024,schmidt2024}. Per-token perplexity cannot be compared across models with different tokenizers \citep{gao2020}. A model's likelihood on reference text need not reflect the properties of the text it generates \citep{meister2021}, and single training runs hide how much results vary with the random seed \citep{dodge2020}. The closest prior benchmark for Romanized Nepali compared three 7–8B models with one run each and per-token perplexity \citep{rimal2026}, leaving open how small models compare and which measures can be trusted to rank them.

We study the adaptation of three 2–3B base models from different families, Llama-3.2-3B, Qwen2.5-3B and Gemma-2-2B, to a Romanized Nepali instruction task. The task is controlled: Romanized text was produced from Devanagari instruction data by deterministic, lowercased ITRANS transliteration, so every word has one spelling (Section~\ref{sec:data}). All models share one QLoRA recipe, a validation split for checkpoint selection and three training seeds. We measure tokenizer fertility, byte-normalized reference likelihood (bits per byte, BPB) and reference-overlap metrics, audit the generated outputs, and compare against zero-shot baselines under two prompt formats. A supplementary comparison with Llama-3.1-8B, whose tokenizer encodes our references identically to that of Llama-3.2-3B, adds a within-family contrast. We ask four questions:

\begin{itemize}[leftmargin=1.6em]
\item \textbf{RQ1.} How do three 2–3B base-model families adapt to Romanized Nepali under one parameter-efficient recipe, and how stable are the outcomes across training seeds?
\item \textbf{RQ2.} Does tokenizer fertility on Romanized Nepali indicate which model adapts best?
\item \textbf{RQ3.} Do reference likelihood and generation overlap agree on how models compare, after adaptation and within one model family when tokenization is identical?
\item \textbf{RQ4.} How do base models behave on this task without adaptation, and how sensitive are zero-shot baselines, and gains measured from them, to the prompt?
\end{itemize}

The paper makes four contributions:

\begin{enumerate}[label=(\arabic*),leftmargin=2.2em]
\item A multi-seed evaluation of three 2–3B base-model families adapted to Romanized Nepali under one QLoRA recipe, reporting both training-seed and test-item uncertainty.
\item Evidence that byte-normalized reference likelihood and generation overlap give different pictures of adapted models, including in a tokenizer-identical within-family comparison.
\item A fertility analysis of four tokenizers on the benchmark references and, under three romanization conventions, on public Nepali text, in which the least-fragmenting tokenizer belonged to the lowest-scoring model and the order of tokenizers was the same under every convention.
\item An analysis of zero-shot behaviour showing that the base models mostly answer in English and that the order of models by gain over zero-shot changes with the prompt.
\end{enumerate}

We document the scripts, hashes and package versions behind every run (Appendix~\ref{app:prov}). The benchmark is a deterministic transliteration of instruction data. It does not test robustness to the spelling variation of naturally typed Romanized Nepali, to mixed scripts or to code-switching, and all measures are automatic, so our conclusions concern adapting and evaluating models on this controlled task. Section~\ref{sec:related} reviews related work, Sections~\ref{sec:data} and \ref{sec:setup} describe the data and experimental setup, Section~\ref{sec:results} reports the results, and Sections~\ref{sec:discussion}–\ref{sec:conclusion} discuss them.

\section{Background and Related Work}\label{sec:related}

\subsection{Romanized Nepali}

Nepali is written in Devanagari, an abugida in which a consonant letter carries an inherent vowel unless it is marked otherwise. The script distinguishes some sounds that spoken Nepali does not: it has long and short \emph{i} and \emph{u}, but spoken Nepali has no vowel-length contrast \citep{khatiwada2009}. Other distinctions are contrastive in speech, such as aspiration and the contrast between dental and retroflex stops \citep{khatiwada2009}. Writers of Romanized Nepali have no agreed way of marking these distinctions in Latin script, so spellings vary across writers \citep{chaudhary2025}. Nepali NLP in Devanagari includes sequence-labelling tools \citep{rai2023}, tokenizer comparisons for language models \citep{luitel2025} and continual pretraining of an 8B model on Nepali \citep{duwal2025}. Work on Romanized Nepali has mostly addressed classification: offensive-language detection in social media, where transliterated text was common \citep{niraula2021}; sentiment analysis after spelling normalization \citep{chaudhary2025}; and sentiment analysis after transliteration back to Devanagari with NepaliXlit \citep{patel2026}.

\subsection{Romanized and Transliterated Text in Language Models}

Romanized text in South Asian languages is common online, and language models handle it unevenly. In real-world triage messages in six Indian languages including Nepali, F1 on romanized text was 5–12 points lower than on native script \citep{khullar2025}. For Sinhala, a benchmark of language models found far higher perplexity on romanized than on native-script text, measured per token across different tokenizers \citep{rajapakse2026}. Romanization can also be used deliberately: RomanSetu uses romanized text as an interface to English-centric models, reducing token fertility by a factor of 2–4 relative to native scripts and matching native-script performance on a range of tasks \citep{husain2024}. In this journal, Baruah et al. \citep{baruah2024} characterized how Romanized Assamese is written in social media and built machine transliteration for it. These studies concern natural romanized text or romanization as a tool. The adaptation of small generative models to produce romanized text, and how such adaptation should be evaluated, has received less attention.

\subsection{Tokenization and Model Performance}

Subword tokenizers built with byte-pair encoding or SentencePiece \citep{sennrich2016,kudo2018} split text that is rare in their training data into more pieces. Fertility, the average number of tokens per word, is the usual measure of this \citep{rust2021}. Tokenized length differs by up to 15 times across languages for the same content \citep{petrov2023}, and for encoder models a dedicated monolingual tokenizer can matter about as much as the amount of pretraining data \citep{rust2021}. Whether fertility predicts downstream performance is less clear. In controlled pretraining, fertility and parity were not always predictive of downstream results \citep{ali2024}; text compression tracked performance, especially for generation tasks and smaller models \citep{goldman2024}; and a tokenizer that minimized token count did not yield better models \citep{schmidt2024}. For Turkish, a study in this journal found a morphological tokenizer competitive with, but not consistently better than, statistical tokenizers \citep{toraman2023}, and for Devanagari Nepali, SentencePiece tokenizers performed better than byte-level BPE on understanding tasks \citep{luitel2025}. In language transfer, extending a model's vocabulary to shorten target-language text was not beneficial at training scales of tens of billions of tokens \citep{zhao2024}. These studies vary the tokenizer under controlled pretraining or transfer. Our question is narrower: whether the fertility of existing tokenizers helps choose among pretrained models to adapt.

\subsection{Parameter-Efficient Adaptation for Low-Resource Languages}

Low-rank adaptation (LoRA) trains small low-rank updates to frozen weights \citep{hu2022}. QLoRA applies it to a 4-bit quantized base model, which reduces memory enough to fine-tune large models on a single GPU \citep{dettmers2023}, and rank-stabilized LoRA (rsLoRA) scales the update by $\alpha$/$\surd$\emph{r} rather than $\alpha$/\emph{r}, which prevents learning from slowing at higher ranks \citep{kalajdzievski2023}. These methods are widely used to adapt open models to low-resource languages. LoRA fine-tuning of Gemma on a translated Alpaca dataset for Marathi lowered automatic scores while manual judgments favoured the tuned model \citep{khade2025}; parameter-efficient fine-tuning of 3–7B models for Bengali hate-speech classification found Llama-3.2-3B best \citep{islam2025}; continual pretraining with 4-bit QLoRA taught Llama 3 8B to generate Nepali, with some forgetting \citep{duwal2025}; and in this journal, Tibetan-LLaMA 2 adapted an open model to Tibetan \citep{sha2025}. All of these concern native-script text.

\subsection{Evaluating Adapted Models}

Generated text is usually compared with references using BLEU \citep{papineni2002}, chrF and chrF++ \citep{popovic2015,popovic2017}, ROUGE \citep{lin2004} or embedding similarity such as BERTScore \citep{zhang2020}. Character-level chrF credits partial matches within words, which suits morphologically rich languages \citep{popovic2015}, but all of these metrics reward closeness to one reference. Likelihood-based measures instead evaluate a model on the reference itself. Per-token perplexity depends on the tokenizer, whereas bits per byte normalizes the total negative log-likelihood by the length of the text in bytes, which makes it comparable across tokenizers \citep{gao2020}. The two kinds of measure can diverge: a model that assigns high probability to human text may still generate text with different statistical properties \citep{meister2021}. Comparisons also need explicit uncertainty. Fine-tuning results vary substantially with the random seed \citep{dodge2020}; differences between systems evaluated on one test set should be assessed with resampling or significance tests \citep{koehn2004,dror2018}; and metric configurations should be reported so that scores can be reproduced \citep{post2018}.

\subsection{The Closest Benchmark}\label{sec:prior}

Rimal and Rimal \citep{rimal2026} fine-tuned Llama-3.1-8B, Mistral-7B-v0.1 and Qwen3-8B on 10,000 Romanized Nepali instruction examples built from the source dataset used here, with half of the instructions machine-translated into English and half transliterated. They trained one QLoRA model with rsLoRA per architecture and reported perplexity, BERTScore, chrF++, ROUGE and BLEU, together with a qualitative case study. All three models adapted, and Qwen3-8B obtained the highest fine-tuned chrF++ (27.47, against 26.97 for Llama-3.1-8B and 23.95 for Mistral-7B-v0.1). They defined adaptation headroom as the fine-tuned minus the zero-shot value of a metric, argued that models with weaker zero-shot performance have more “representational plasticity”, and attributed Llama's difficulty to “its Tiktoken tokenizer over-fragmenting Romanized Nepali into low-frequency subword units”, leaving the measurement of fertility to future work. Their study leaves open how smaller models compare, how much results vary across training runs, whether directly measured fertility supports the over-fragmentation account, and how likelihood compares across models with different tokenizers. We are not aware of a study that measures tokenizer fertility, byte-normalized reference likelihood and generation overlap together for small models adapted to Romanized Nepali, with seed-level uncertainty. Our recipe follows theirs (Section~\ref{sec:recipe}), but model scale, data processing, checkpoint selection, decoding and the likelihood measure differ, so we compare findings, not scores.

\section{Data}\label{sec:data}

\subsection{Source, Filtering and Splits}\label{sec:source}

The source is the Nepali instruction dataset Saugatkafley/alpaca-nepali-sft on the Hugging Face Hub, with 52,005 examples whose instruction, input and output fields are in Devanagari. Its dataset card states no licence and gives little information on provenance, but the data are a Nepali translation of the Stanford Alpaca data \citep{taori2023}, which is released under CC~BY-NC~4.0. The type A instructions of our test split (Section~\ref{sec:formats}), translated back into English, match Alpaca instructions closely: 29.8\% are identical to one of its 52,002 instructions, and 92.2\% have a normalized edit similarity of at least 0.8 to the nearest one. We follow the data recipe of Rimal and Rimal \citep{rimal2026} with our own transliteration choices. We removed examples that still contained Devanagari characters after transliteration and examples whose outputs were shorter than three words, which left 48,106 eligible examples; examples with an empty input field were kept. From these we drew 10,500 examples (sampling seed 42) and split them into 9,000 training, 500 validation and 1,000 test examples, with the two instruction types (Section~\ref{sec:formats}) balanced 50/50 in every split. The 500-example validation split, which the prior recipe does not describe, accounts for the difference from its 10,000 examples.

Empty inputs occur in 5,576 training, 314 validation and 660 test examples. Split files are identified by SHA-256 prefixes (train 9ed86da7\ldots{}, validation 3f7f6854\ldots{}, test d656353e\ldots{}; full hashes in the code repository). We found no exact instruction overlap between splits. One output is shared between the training and validation splits and three between the training and test splits, each of three or more words, and we kept them. We also kept two artifacts of the source data: placeholder strings meaning “no input” in 114 training, 5 validation and 14 test examples, and zero-width-space characters in 113, 11 and 12 examples.

\subsection{Romanization}\label{sec:roman}

Romanized text was produced by ITRANS transliteration of the Devanagari fields, followed by lowercasing. Rimal and Rimal \citep{rimal2026} describe their transliteration as using an Indic transliteration library with phonetic variation permitted, so the scheme is our choice. Table~\ref{tab:roman} shows how it treats some common words. It has three properties that matter for interpreting the results.

First, it is deterministic: every Devanagari word has exactly one romanized spelling, so the benchmark contains none of the spelling variation of naturally typed Romanized Nepali.

Second, lowercasing removes the case distinctions that ITRANS uses to encode Devanagari letters, which merges 11 groups of letters: eight consonant groups (\dev{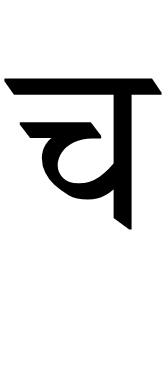}/\dev{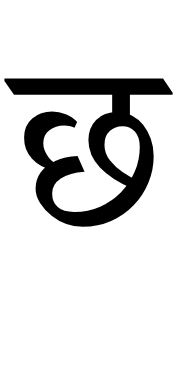}, \dev{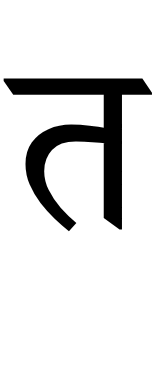}/\dev{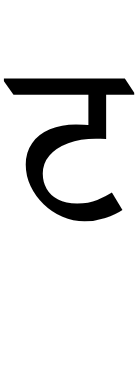}, \dev{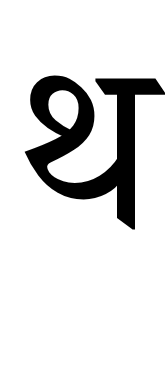}/\dev{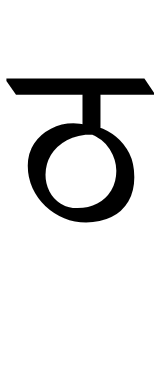}, \dev{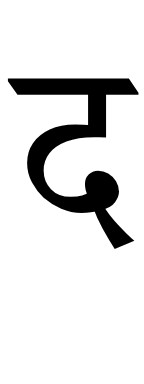}/\dev{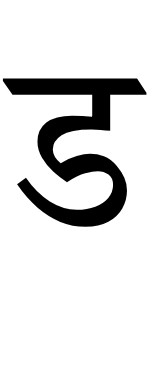}, \dev{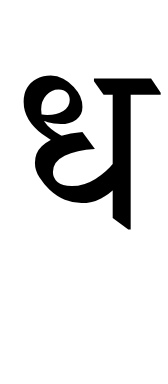}/\dev{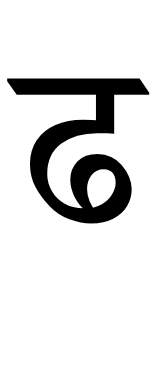}, \dev{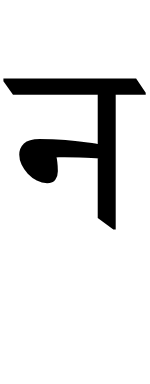}/\dev{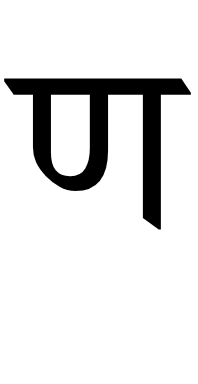}, \dev{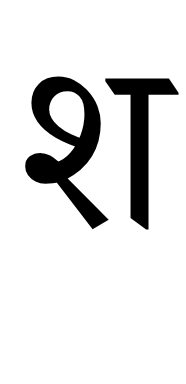}/\dev{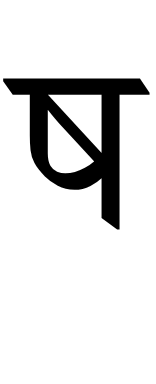} and \dev{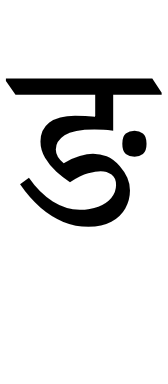}/\dev{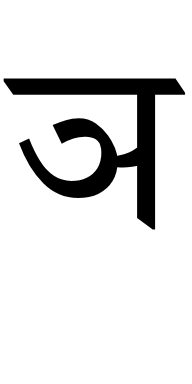}) and three vowel groups (\dev{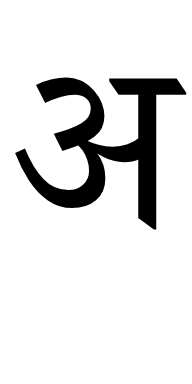}/\dev{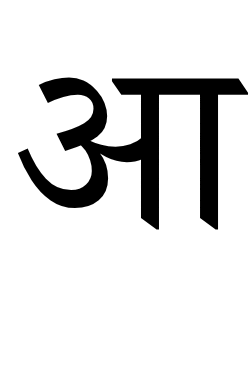}, \dev{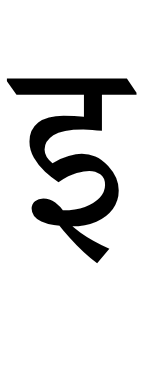}/\dev{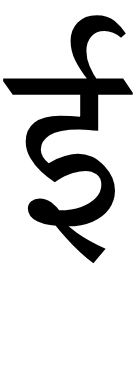} and \dev{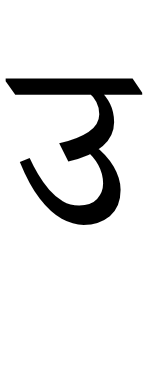}/\dev{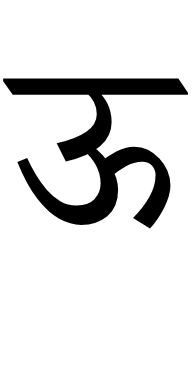}, with their vowel signs; Appendix~\ref{app:roman}). Some merged pairs are contrastive in spoken Nepali, such as unaspirated and aspirated affricates (\dev{deva/d12.png}/\dev{deva/d13.png}) and dental and retroflex stops (\dev{deva/d21.png}/\dev{deva/d15.png}, \dev{deva/d25.png}/\dev{deva/d18.png}), whereas the long and short vowels are distinguished only in the script \citep{khatiwada2009}. To estimate how much text the merger affects, we transliterated a public sample of 1,233 Nepali sentences (Section~\ref{sec:res-fert} and Appendix~\ref{app:roman}). In this sample, 58.8\% of running words contain at least one uppercase ITRANS character, so lowercasing changes the spelling of most words, and 3.8\% of the resulting word types merge two or more distinct ITRANS spellings. These merged types are frequent: they cover 13.8\% of running words, as with \dev{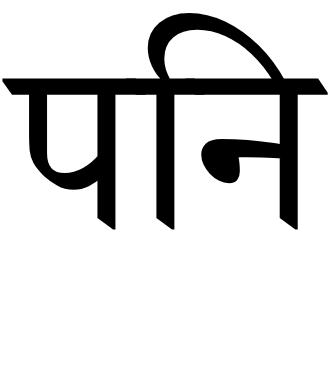} and \dev{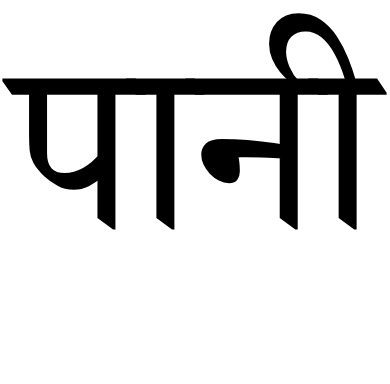} (both \emph{pani}) or \dev{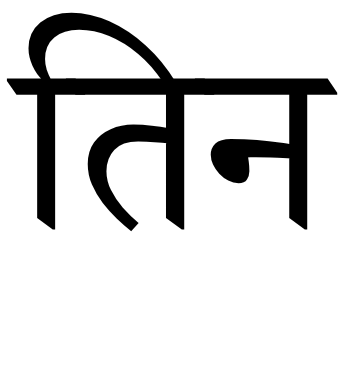}, \dev{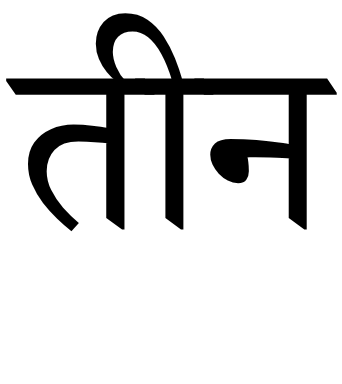} and \dev{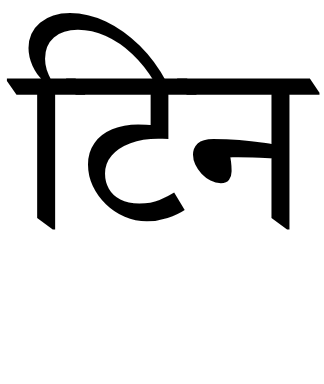} (all \emph{tina}).

Third, the scheme keeps the inherent vowel at the end of a word, which conventional romanizations often drop: \dev{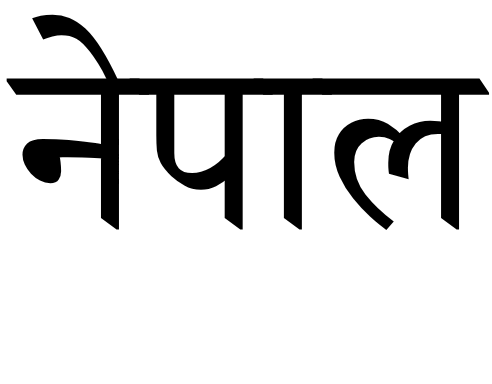} is conventionally written \emph{Nepal}, but the scheme gives \emph{nepala}. In the public sample, 36.2\% of running words end in an inherent vowel, and 29.4\% are words of two or more aksharas (orthographic syllables) that do so. For the fertility analysis we therefore also use two alternative conventions: case-preserving ITRANS (R1) and the benchmark scheme without the final inherent vowel in words of two or more aksharas (R3), a simple approximation of informal spelling. The benchmark scheme itself is R2.

\begin{table}[tbp]
\centering
\fontsize{8.5}{10}\selectfont
\setlength{\tabcolsep}{3pt}
\caption{Example words under the three romanization conventions. R2 (lowercased ITRANS) is the benchmark scheme; R1 and R3 are used only in the fertility analysis (Section~\ref{sec:res-fert}). Under R2, \dev{deva/d29.png} and \dev{deva/d30.png}, \dev{deva/d23.png} and \dev{deva/d16.png}, and \dev{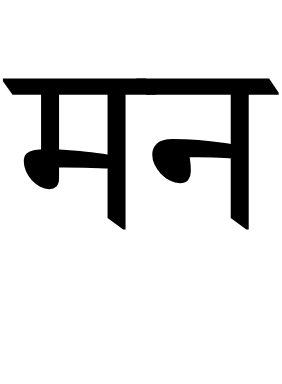} and \dev{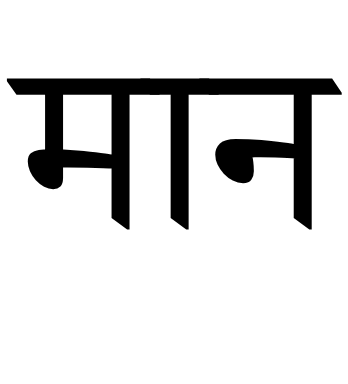} become identical.}\label{tab:roman}
\begin{tabular}{@{}L{83.9pt}L{82.7pt}L{91.1pt}L{92.6pt}L{89.5pt}@{}}
\toprule
\textbf{Devanagari} & \textbf{Gloss} & \textbf{R1: ITRANS} & \textbf{R2: lowercased} & \textbf{R3: R2, final vowel dropped} \\
\midrule
\dev{deva/d28.png} & Nepal & \texttt{nepAla} & \texttt{nepala} & \texttt{nepal} \\
\dev{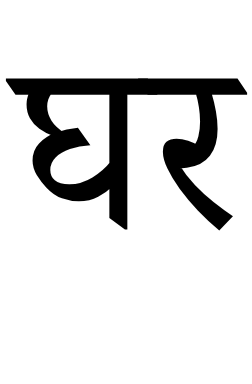} & house & \texttt{ghara} & \texttt{ghara} & \texttt{ghar} \\
\dev{deva/d13.png} & is & \texttt{Cha} & \texttt{cha} & \texttt{cha} \\
\dev{deva/d29.png} & also & \texttt{pani} & \texttt{pani} & \texttt{pani} \\
\dev{deva/d30.png} & water & \texttt{pAnI} & \texttt{pani} & \texttt{pani} \\
\dev{deva/d23.png} & three & \texttt{tIna} & \texttt{tina} & \texttt{tin} \\
\dev{deva/d16.png} & tin, can & \texttt{Tina} & \texttt{tina} & \texttt{tin} \\
\dev{deva/d32.png} & mind & \texttt{mana} & \texttt{mana} & \texttt{man} \\
\dev{deva/d34.png} & respect & \texttt{mAna} & \texttt{mana} & \texttt{man} \\
\dev{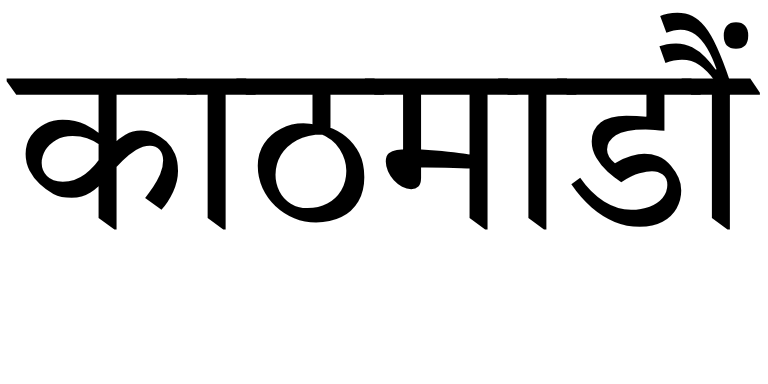} & Kathmandu & \texttt{kAThamADauM} & \texttt{kathamadaum} & \texttt{kathamadaum} \\
\dev{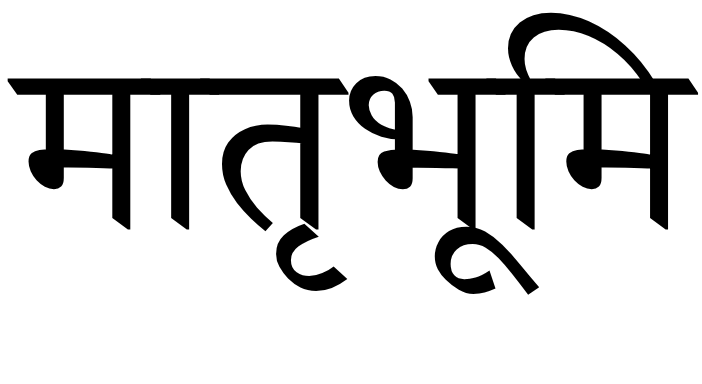} & motherland & \texttt{mAtRRibhUmi} & \texttt{matrribhumi} & \texttt{matrribhumi} \\
\bottomrule
\end{tabular}
\end{table}

\subsection{Instruction Formats}\label{sec:formats}

Following the prior recipe \citep{rimal2026}, each example has one of two formats. In type A, the instruction was machine-translated from Nepali into English and the input and output were transliterated; in type B, all three fields were transliterated. Type A instructions have therefore been translated twice, from English into Nepali in the source dataset and back into English here, and back-translation sometimes changes the meaning: one test instruction asks why chloroplasts are found “in plant cells and animal cells”, where the Alpaca original has “in plant cells and not animal cells”. Prompts follow the Alpaca template \citep{taori2023}, with its no-input variant for examples without an input, and the target is always the Romanized Nepali output.

\subsection{Length}

Training sequences (prompt plus response) were limited to 512 tokens, which truncated 178 (Qwen2.5-3B), 164 (Llama-3.2-3B) and 127 (Gemma-2-2B) of the 9,000 training examples (1.4–2.0\%) and 6, 6 and 5 of the 500 validation examples. In the test set, 19, 17 and 14 examples exceed 512 tokens under the same tokenizers, but no generation prompt does (Section~\ref{sec:decoding}). The 1,000 test references contain 41,388 whitespace-delimited words, 41.4 per response on average, and 124 (Qwen2.5-3B), 117 (Llama-3.2-3B) and 88 (Gemma-2-2B) of them are longer than the 256-token generation limit.

\section{Experimental Setup}\label{sec:setup}

\subsection{Models}\label{sec:models}

All models are base checkpoints that were not instruction-tuned: meta-llama/Llama-3.2-3B \citep{meta2024b}, Qwen/Qwen2.5-3B \citep{qwen2024} and google/gemma-2-2b \citep{gemma2024} in the main comparison, and meta-llama/Llama-3.1-8B \citep{grattafiori2024,meta2024a} in the supplementary within-family comparison (Table~\ref{tab:models}; revisions in Appendix~\ref{app:prov}). The three primary models differ in more than the tokenizer. Gemma-2-2B has about 2.0 billion non-embedding parameters against 2.8 billion for the other two, saw fewer pretraining tokens, and puts 22.6\% of its parameters into the embeddings of its 256k-entry vocabulary. Llama-3.2-3B and Gemma-2-2B were trained with knowledge distillation from larger models \citep{meta2024b,gemma2024}.

\begin{table}[tbp]
\centering
\fontsize{8.5}{10}\selectfont
\setlength{\tabcolsep}{3pt}
\caption{Models. Parameter counts are computed from the published model configurations; the embedding share of Llama-3.1-8B includes its separate output matrix. Vocabulary, pretraining tokens and licences are from the model reports and cards \citep{gemma2024,grattafiori2024,meta2024a,meta2024b,qwen2024}. The LoRA adapter sizes implied by these configurations match the trainable-parameter counts in Section~\ref{sec:recipe}.}\label{tab:models}
\begin{tabular}{@{}L{52.1pt}R{26.4pt}R{57.7pt}R{40.8pt}R{44.9pt}L{60.4pt}R{49.1pt}L{90.4pt}@{}}
\toprule
\textbf{Model} & \textbf{Total params} & \textbf{Non-embedding params} & \textbf{Embedding share} & \textbf{Vocabulary} & \textbf{Tokenizer} & \textbf{Pretraining tokens} & \textbf{Licence} \\
\midrule
Llama-3.2-3B & 3.21B & 2.82B & 12.3\% & 128,256 & byte-level BPE & up to 9T & Llama 3.2 Community License \\
Qwen2.5-3B & 3.09B & 2.77B & 10.1\% & 151,643 & byte-level BPE & 18T & Qwen Research License \\
Gemma-2-2B & 2.61B & 2.02B & 22.6\% & 256k & SentencePiece & 2T & Gemma Terms of Use \\
Llama-3.1-8B & 8.03B & 6.98B & 13.1\% & 128,256 & byte-level BPE & \textasciitilde{}15T & Llama 3.1 Community License \\
\bottomrule
\end{tabular}
\end{table}

\subsection{Adaptation Recipe}\label{sec:recipe}

Each model is adapted with QLoRA \citep{dettmers2023} through Unsloth \citep{unsloth2023}. Base weights are loaded in 4-bit NF4 with double quantization and fp16 compute, and rank-stabilized LoRA adapters \citep{kalajdzievski2023} (\emph{r}~=~32, $\alpha$~=~64, dropout 0, no bias) are trained on the attention (\texttt{q\_proj}, \texttt{k\_proj}, \texttt{v\_proj}, \texttt{o\_proj}) and feed-forward (\texttt{gate\_proj}, \texttt{up\_proj}, \texttt{down\_proj}) projections. With rank-stabilized scaling the adapter output is multiplied by $\alpha$/$\surd$\emph{r}~=~64/$\surd$32~$\approx$~11.3, which is 5.7 times the standard LoRA factor $\alpha$/\emph{r}~=~2 \citep{hu2022}. The adapters have 59.9M trainable parameters for Qwen2.5-3B, 48.6M for Llama-3.2-3B, 41.5M for Gemma-2-2B and 83.9M for Llama-3.1-8B (1.0–1.9\% of the total). Unsloth substitutes its own pre-quantized 4-bit checkpoints where they exist (the Llama-3.2-3B zero-shot log shows \texttt{unsloth/Llama-3.2-3B-unsloth-bnb-4bit}) and leaves some modules unquantized, depending on the model: \texttt{lm\_head} and the MLPs of layers 1, 2, 3, 28 and 30 for Qwen2.5-3B, \texttt{lm\_head} and the layer-1 MLP for Llama-3.2-3B, and none for Gemma-2-2B (measured with Unsloth 2026.11.2; runs under other versions may differ).

Training uses a learning rate of 1~$\times$~10\textsuperscript{$-$4} with cosine decay and 200 warm-up steps, 8-bit AdamW, three epochs (3,375 optimizer steps), a maximum sequence length of 512 tokens without packing, fp16, and an effective batch size of 8 (2 per device $\times$ 4 accumulation steps) on a single NVIDIA T4 GPU. Each model is trained with seeds 42, 43 and 44. The recipe follows Rimal and Rimal \citep{rimal2026} except for the batch size and the validation split. It was shared across models and tuned for none of them, so it may suit some models better than others. For Gemma-2-2B, Unsloth compilation was disabled (\texttt{UNSLOTH\_COMPILE\_DISABLE=1}) in training and evaluation for all three seeds.

\subsection{Checkpoint Selection and Divergence}\label{sec:ckpt}

We evaluate on the validation split every 100 steps and keep the checkpoint with the lowest validation loss; training is not stopped early. Appendix~\ref{app:prov} lists every training run and its status. A second Qwen2.5-3B seed-42 run, launched because of a Kaggle session limit and to keep package versions the same across seeds, diverged: its validation loss jumped from 1.46 at step 1200 to 6.99 at step 1300, 4.8 times the best earlier value, before recovering. After it, we defined a run as diverged if any validation loss after step 500 exceeded 1.5 times the best earlier validation loss, and we report the first non-diverged run for each seed. The rule uses validation loss only, but it was fixed after the diverged run's test scores had been seen, so it is a post hoc diagnostic rather than a pre-specified exclusion rule. No other run met it. The maximum post-step-500 ratio was 1.14–1.15 for Gemma-2-2B and 1.107–1.115 for Llama-3.1-8B; the retained records do not give the maxima for the other runs, and they give the selected step only for Qwen2.5-3B seeds 43 and 44 (step 2200 in both). Llama-3.2-3B seed 43 logged one non-finite gradient norm at step 2150, and training continued normally.

\subsection{Zero-Shot Conditions}\label{sec:zeroshot}

The zero-shot conditions use the base checkpoints without adapters under two prompts: (i) the Alpaca template used in fine-tuning, with its no-input variant where there is no input, and (ii) a bare prompt consisting of the instruction followed by the input on a new line. Each condition is a single deterministic run. The base models were not instructed to answer in Romanized Nepali, so these conditions measure what the models do with the task as given, not their best attainable zero-shot performance.

\subsection{Decoding and Likelihood}\label{sec:decoding}

All conditions share decoding and scoring. We decode greedily with at most 256 new tokens and a repetition penalty of 1.1, in fp16 with batch size 8 and left padding; for fine-tuned models the adapter is applied to the 4-bit base without merging. Prompts are truncated to 512 tokens, a limit that no test prompt reached.

We also score each reference response given its prompt by teacher forcing. For test item \emph{i}, let NLL\emph{\textsubscript{i}} be the summed negative log-likelihood, in nats, of the reference response tokens, excluding the end-of-sequence token, and let \emph{B}\emph{\textsubscript{i}} be the length of the reference in UTF-8 bytes. Prompt and response together are capped at 1,000 tokens, which truncated one item (id 289) under the Qwen and Llama tokenizers. Over the \emph{N}~=~1,000 test items, bits per byte is

\begin{equation}\label{eq:bpb}
\mathrm{BPB} = \frac{\sum_{i=1}^{N} \mathrm{NLL}_i}{\ln 2 \, \sum_{i=1}^{N} B_i}
\end{equation}

Because the denominator counts bytes rather than tokens, BPB can be compared across models with different tokenizers \citep{gao2020}; lower values mean that the model assigns higher probability to the references. We also compute reference token perplexity, the exponential of the pooled NLL per response token, but compare it only within a model. BPB under the two zero-shot prompts is not compared, because the prompts condition the model differently.

\subsection{Generation Metrics}\label{sec:metrics}

We report corpus-level BLEU \citep{papineni2002} and chrF++ (character n-grams with word order 2) \citep{popovic2017}, computed with sacreBLEU 2.6.0 \citep{post2018} on a 0–100 scale (signatures in Appendix~\ref{app:prov}); ROUGE-1, ROUGE-2 and ROUGE-L F-scores \citep{lin2004}, computed with the rouge-score package without stemming; and BERTScore F1 \citep{zhang2020} with multilingual BERT (bert-base-multilingual-cased \citep{devlin2019}; bert-score 0.3.13, no baseline rescaling), averaged over items. Each test item has one reference. The BERTScore encoder was pretrained on text in which Nepali appears in Devanagari and has not been validated for Romanized Nepali, so we treat BERTScore as a supplementary similarity score. Empty generations are kept in every metric and score 0 in BERTScore; for runs with empty generations we also give the BERTScore mean over non-empty outputs as a descriptive supplement. Recomputing chrF++, BLEU, ROUGE and BPB from the saved per-item outputs reproduces the values of the evaluation scripts for every run (Appendix~\ref{app:seeds}); BERTScore could not be recomputed in our analysis environment.

In our environment (bert-score 0.3.13 with Transformers 5.17.0), the evaluation script failed at the BERTScore step for zero-shot runs that contained at least one empty generation, after generation and likelihood scoring had finished. For these runs we rebuilt all text metrics from the saved outputs with a separate script that reuses the saved NLL values and bootstrap settings and does not regenerate. On the Qwen2.5-3B Alpaca run, which had no empty outputs, this script reproduced BLEU, chrF++, ROUGE and BPB exactly and BERTScore to within 10\textsuperscript{$-$7}.

\subsection{Uncertainty}\label{sec:uncert}

We report two kinds of uncertainty. Variation across training seeds is given as the mean and sample standard deviation (SD) over three seeds; with three runs the SD itself is imprecise, since a 95\% interval for the true SD runs from 0.52 to 6.3 times the sample SD. Test-set sampling variation is estimated by resampling the 1,000 test items from the saved outputs, which hold each generation, its reference and the reference NLL. Per-run 95\% percentile intervals use 2,000 resamples (seed 0). To compare two models, we resample the items once per replicate (2,000 replicates, seed 0), recompute corpus chrF++, BLEU and pooled BPB for every seed of both models on the resampled items, average over each model's seeds and take the difference. These paired item-level intervals condition on the trained seeds. As a complement that conditions on the test set instead, Table~\ref{tab:diffs} also gives seed-level intervals (Welch's \emph{t} over the three per-seed scores of each model). Fertility intervals use 1,000 bootstrap resamples over references (seed 0) and paired fertility differences 2,000 resamples (seed 1). The evaluation scripts' own per-run intervals used 200 resamples (Python random seed 0); the intervals reported here replace them. We do not interpret differences that neither kind of interval resolves.

\subsection{Output Audit}\label{sec:audit}

One script (SHA-256 69b88899\ldots{}) audits every saved output. It flags (i) English-like output, when English stopwords make up at least 15\% of alphabetic tokens, using a fixed 111-word list screened against the reference responses; (ii) input copying, when the first non-empty line, of at least three words, reproduces a non-template line of the prompt; and (iii) repetition loops, when at least half of the non-empty lines repeat earlier lines, for outputs of at least three lines. The English and loop checks apply only to outputs long enough to evaluate, so their denominators vary. The same definitions apply to all 20 evaluation runs. As a descriptive baseline, the audit of the 1,000 reference responses flags none of 922 evaluable responses as English-like, 10 as input copies and 1 of 181 evaluable as a repetition loop. Because the stopword list was screened against these references, the baseline does not validate the English flag independently. The flags are surface heuristics, not judgments of correctness: English words can be appropriate in code or explanations, and repeated lines can occur in valid structured answers.

\subsection{Software and Provenance}\label{sec:software}

The primary fine-tuned evaluations used evaluation script v2.2 and the Llama-3.1-8B evaluations v2.4 (51e1bd41\ldots{}). The fine-tuned code path (prompt template, generation settings, likelihood scoring and metrics) is identical in the two versions; v2.4 adds the two zero-shot prompt modes, saves generations before computing metrics, and corrects a field that left the v2.2 metric files recording their script hash as unknown. The zero-shot runs of the primary models used one script (33e10eac\ldots{}) with per-run overrides; Appendix~\ref{app:prov} lists which patched copies have recorded hashes. Package versions differed across sessions (Appendix~\ref{app:prov}). To check that this does not change results, we re-evaluated Qwen2.5-3B seed 43 in a later environment (Unsloth 2026.11.2, Transformers 5.17.0, PEFT 0.21.2 and PyTorch built for CUDA 13.0): all 1,000 generations, all per-item NLL values and every metric were identical. This check covers one adapter. The training and evaluation code, the re-analysis of the saved outputs (Sections~\ref{sec:res-zs}–\ref{sec:res-8b} and Appendix~\ref{app:seeds}), the tokenizer and romanization analysis (Section~\ref{sec:res-fert} and Appendix~\ref{app:roman}) and the figure code were written with the assistance of an AI model (Acknowledgments).

\section{Results}\label{sec:results}

\subsection{Tokenizer Fertility on Romanized Nepali}\label{sec:res-fert}

On the 1,000 test references (41,388 words), each encoded whole without special tokens, the Gemma-2-2B tokenizer has the lowest fertility, 2.717 tokens per word, followed by Llama-3.2-3B (2.938) and Qwen2.5-3B (2.997) (Table~\ref{tab:fert}). Medians and tokens per character give the same order, and paired bootstrap differences confirm it: Llama exceeds Gemma by 0.221 [0.206, 0.235], Qwen exceeds Llama by 0.059 [0.051, 0.067], and Qwen exceeds Gemma by 0.280 [0.268, 0.292]. Per reference, Llama's tokenizer needed fewer tokens per word than Gemma's for only 9.1\% of the references, and Qwen's for only 4.5\%. The Mistral-7B-v0.1 tokenizer, which we did not fine-tune, splits words more (3.295). The Llama-3.1-8B and Llama-3.2-3B tokenizers produced identical token IDs for all 1,000 references, as did the Qwen2.5-3B and Qwen3-8B tokenizers. Among the trained models the range is narrow: 2.72–3.00 tokens per word, a 10\% spread.

\begin{table}[tbp]
\centering
\fontsize{8.5}{10}\selectfont
\setlength{\tabcolsep}{3pt}
\caption{Tokenizer fertility on the 1,000 benchmark references (41,388 words) beside the three-seed fine-tuned results. Fertility is the mean over references of tokens per whitespace-delimited word, with 95\% bootstrap intervals (1,000 resamples). Higher chrF++ and lower BPB are better. Mistral-7B-v0.1 was not trained here.}\label{tab:fert}
\begin{tabular}{@{}L{143.4pt}R{94.5pt}R{39.0pt}R{41.3pt}R{58.2pt}R{57.4pt}@{}}
\toprule
\textbf{Tokenizer} & \textbf{Tokens per word [95\% CI]} & \textbf{Median} & \textbf{Tokens per char} & \textbf{Fine-tuned chrF++} & \textbf{Fine-tuned BPB} \\
\midrule
Gemma-2-2B & 2.717 [2.689, 2.747] & 2.677 & 0.341 & 29.02 & 0.842 \\
Llama-3.2-3B (= Llama-3.1-8B) & 2.938 [2.908, 2.968] & 2.928 & 0.374 & 30.51 & 0.779 \\
Qwen2.5-3B (= Qwen3-8B) & 2.997 [2.965, 3.029] & 3.000 & 0.380 & 30.61 & 0.821 \\
Mistral-7B-v0.1 & 3.295 [3.262, 3.329] & 3.276 & 0.419 & – & – \\
\bottomrule
\end{tabular}
\end{table}

Because the benchmark references are themselves in one romanization scheme, we also measured fertility on public Nepali text under the three conventions of Section~\ref{sec:roman} (Table~\ref{tab:conv}; details in Appendix~\ref{app:roman}). The sample comprises 1,233 sentences: 1,146 from the Mozilla Common Voice Nepali text corpus \citep{ardila2020} and 87 lines of the Universal Declaration of Human Rights in Nepali distributed with NLTK \citep{bird2009}. Under the benchmark scheme (R2), tokens per character on this sample (0.339–0.420) are close to those on the benchmark references (0.341–0.419), but fertility is lower (2.13–2.65), because the sample's words are shorter (about 6.4 against 7.9 characters per word, counting the following space). The tokenizer files used for this sample reproduce the Hugging Face token counts exactly on all 1,000 benchmark references (Appendix~\ref{app:roman}).

\begin{table}[tbp]
\centering
\fontsize{8.5}{10}\selectfont
\setlength{\tabcolsep}{3pt}
\caption{Fertility on a public sample of 1,233 Nepali sentences under three romanization conventions: R1, case-preserving ITRANS; R2, lowercased ITRANS (the benchmark scheme); R3, R2 without the final inherent vowel in words of two or more aksharas. Word shares are over 8,840 running words under R2. Bootstrap intervals have a half-width of 0.03–0.04 under every convention.}\label{tab:conv}
\begin{tabular}{@{}L{76.4pt}R{37.0pt}R{106.2pt}R{37.0pt}R{74.3pt}R{49.1pt}R{47.8pt}@{}}
\toprule
\textbf{Tokenizer} & \textbf{R1} & \textbf{R2 [95\% CI]} & \textbf{R3} & \textbf{R2 single-token words} & \textbf{R2 words of $\geq$3 tokens} & \textbf{R2 tokens per char} \\
\midrule
Gemma-2 & 2.493 & 2.135 [2.107, 2.165] & 2.069 & 30.9\% & 31.6\% & 0.339 \\
Llama 3 & 2.712 & 2.358 [2.332, 2.385] & 2.264 & 22.4\% & 39.7\% & 0.377 \\
Qwen2.5 & 2.733 & 2.387 [2.359, 2.414] & 2.294 & 22.2\% & 40.9\% & 0.381 \\
Mistral v0.1 & 3.053 & 2.648 [2.612, 2.683] & 2.536 & 19.0\% & 49.7\% & 0.420 \\
\bottomrule
\end{tabular}
\end{table}

The order Gemma < Llama < Qwen < Mistral held in all 2,000 paired bootstrap resamples under every convention. The convention, however, changed fertility about as much as the choice of tokenizer did. Lowercasing (R1 to R2) lowered fertility by 12.7–14.4\%, or 0.35–0.41 tokens per word, which approaches the full spread across the four tokenizers under R2 (0.51) and is more than ten times the Llama–Qwen difference (0.029). Dropping the final inherent vowel (R2 to R3) lowered it by a further 3.1–4.2\%. Gemma's lower fertility comes from keeping more words whole: under R2, 30.9\% of running words are a single Gemma token, against 19.0–22.4\% for the other tokenizers, and 31.6\% are split into three or more tokens, against 39.7–49.7\%.

Fertility did not correspond to the fine-tuned results (Table~\ref{tab:fert}). Gemma-2-2B, whose tokenizer splits words least, has the lowest overlap scores and the highest BPB of the three models (Section~\ref{sec:res-adapt}), and Llama-3.2-3B has the lowest BPB although its tokenizer splits words more than Gemma's. With three trained models that also differ in size, data and architecture (Table~\ref{tab:models}), this is a counterexample to choosing models by fertility, not an estimate of its effect.

\subsection{Zero-Shot Behaviour}\label{sec:res-zs}

Without adaptation, all three base models score far below their fine-tuned versions (Table~\ref{tab:zs}). chrF++ is 10.1–13.7, against 29.0–30.6 after fine-tuning, and BPB under the Alpaca prompt is 2.23–2.41, against 0.78–0.84. Most zero-shot outputs are not Romanized Nepali at all: 543–923 of 1,000 outputs per condition (54.3–92.3\%) are flagged as English-like, against none of the evaluable references (Table~\ref{tab:audit}).

\begin{table}[tbp]
\centering
\fontsize{8.5}{10}\selectfont
\setlength{\tabcolsep}{3pt}
\caption{Zero-shot results on the 1,000 test items (one deterministic run per condition). Empty generations are kept and score 0 in BERTScore; the mean over non-empty outputs is in parentheses. BPB is comparable across models within a prompt, not across prompts.}\label{tab:zs}
\begin{tabular}{@{}L{90.6pt}L{50.6pt}R{52.8pt}R{46.7pt}R{88.7pt}R{55.7pt}R{42.8pt}@{}}
\toprule
\textbf{Model} & \textbf{Prompt} & \textbf{chrF++} & \textbf{BLEU} & \textbf{BERTScore F1} & \textbf{Empty outputs} & \textbf{BPB} \\
\midrule
Qwen2.5-3B & Alpaca & 13.65 & 0.64 & 0.545 & 0 & 2.373 \\
Qwen2.5-3B & bare & 11.45 & 0.43 & 0.492 (0.541) & 91 & 2.404 \\
Llama-3.2-3B & Alpaca & 12.51 & 1.01 & 0.587 & 4 & 2.225 \\
Llama-3.2-3B & bare & 12.65 & 0.38 & 0.536 & 9 & 2.276 \\
Gemma-2-2B & Alpaca & 10.13 & 0.41 & 0.539 & 3 & 2.415 \\
Gemma-2-2B & bare & 10.96 & 0.12 & 0.526 & 2 & 2.466 \\
\bottomrule
\end{tabular}
\end{table}

The effect of the prompt depends on the model. For Qwen2.5-3B the Alpaca prompt gives higher chrF++ (13.65 [13.32, 13.98] against 11.45 [11.04, 11.86]) and BERTScore, and a higher BLEU point estimate with overlapping intervals (0.64 [0.52, 0.77] against 0.43 [0.29, 0.60]). For Llama-3.2-3B, chrF++ is similar under the two prompts (12.51 and 12.65), while BLEU and BERTScore are higher with the Alpaca template. Gemma-2-2B has lower chrF++ under the Alpaca prompt (10.13 against 10.96) but higher BLEU and BERTScore. We report these as point estimates and do not claim that the smaller differences are reliable. Qwen2.5-3B's bare-prompt BERTScore of 0.492 includes 91 empty generations scored as 0; over its 909 non-empty outputs the mean is 0.541, close to its Alpaca value. Likelihood already orders the base models: under both prompts, BPB is lowest for Llama-3.2-3B and highest for Gemma-2-2B, and every paired difference is resolved.

The failures differ by model and prompt (Table~\ref{tab:audit}). Llama-3.2-3B copied the prompt in 135 outputs under the Alpaca template, mostly on type B items (24.2\% of type B outputs), and produced 110 repetition loops under the bare prompt (16.4\% of type B outputs). Qwen2.5-3B produced 91 empty generations and 29 loops under the bare prompt, against none and 2 under the Alpaca template. Gemma-2-2B produced few copying or repetition flags under either prompt, and the most English-like output across the two prompts (830 and 877). Where the audit breaks failures down by type, they concentrate on type B prompts, whose instructions are themselves Romanized. After fine-tuning, the nine runs had 0–2 English-like flags, 4–20 input copies (0.4–2.0\%), 0–2 loops and no empty generations, rates comparable to those of the references. The flagged fine-tuned outputs that we inspected mostly answered prompts whose references are themselves structured or code (an HTML table, SQL, JavaScript or numbered lists) or contain quoted English words.

\begin{table}[tbp]
\centering
\fontsize{8.5}{10}\selectfont
\setlength{\tabcolsep}{3pt}
\caption{Output audit (heuristic flags, not quality judgments). Counts are out of 1,000 outputs unless an evaluable denominator is shown; the English and loop checks apply only to outputs long enough to evaluate.}\label{tab:audit}
\begin{tabular}{@{}L{144.3pt}R{94.3pt}R{52.4pt}R{96.4pt}R{52.4pt}@{}}
\toprule
\textbf{Condition} & \textbf{English-like} & \textbf{Input copy} & \textbf{Repetition loop} & \textbf{Empty} \\
\midrule
Reference responses & 0 / 922 & 10 & 1 / 181 & 0 \\
Fine-tuned (9 runs) & 0–2 / 909–931 & 4–20 & 0–2 / 175–210 & 0 \\
Qwen2.5-3B, Alpaca & 923 & 2 & 2 & 0 \\
Qwen2.5-3B, bare & 608 & 3 & 29 & 91 \\
Llama-3.2-3B, Alpaca & 543 & 135 & 24 & 4 \\
Llama-3.2-3B, bare & 641 & 22 & 110 / 794 & 9 \\
Gemma-2-2B, Alpaca & 830 & 3 & 0 & 3 \\
Gemma-2-2B, bare & 877 & 1 & 6 & 2 \\
\bottomrule
\end{tabular}
\end{table}

\subsection{Adaptation across Model Families}\label{sec:res-adapt}

All three models adapt strongly (Table~\ref{tab:ft}). Mean chrF++ rises to 29.02–30.61 and BPB falls to 0.779–0.842, with no empty outputs and few audit flags. Qwen2.5-3B and Llama-3.2-3B cannot be separated on any generation metric: their mean chrF++ differs by 0.09 points, BLEU by 0.04 and ROUGE-1 by 0.001, and their seed ranges overlap (chrF++ 29.53–31.42 and 29.97–31.38). Gemma-2-2B is lower on every overlap metric. Its BLEU separates cleanly, with all three seeds (4.51–4.99) below every Qwen and Llama seed (5.61 or above), whereas its chrF++ overlaps: its best seed (29.73) exceeds the lowest Qwen seed (29.53). BERTScore differs little between models (0.750–0.757). All three models score higher on type A than on type B, by 2.4–2.8 chrF++ points (type A 30.25–31.92, type B 27.83–29.40), and BPB is lower on type A in every model.

\begin{table}[tbp]
\centering
\fontsize{8.5}{10}\selectfont
\setlength{\tabcolsep}{3pt}
\caption{Fine-tuned results on the 1,000 test items: mean $\pm$ sample SD over seeds 42, 43 and 44. Higher is better except for BPB. Qwen2.5-3B seed 42 is the first non-diverged run (Section~\ref{sec:ckpt}).}\label{tab:ft}
\begin{tabular}{@{}L{58.4pt}R{49.6pt}R{45.1pt}R{53.7pt}R{53.7pt}R{53.7pt}R{53.7pt}R{53.7pt}@{}}
\toprule
\textbf{Model} & \textbf{chrF++} & \textbf{BLEU} & \textbf{ROUGE-1} & \textbf{ROUGE-2} & \textbf{ROUGE-L} & \textbf{BERTScore F1} & \textbf{BPB} \\
\midrule
Qwen2.5-3B & 30.61 $\pm$ 0.97 & 5.86 $\pm$ 0.26 & 0.252 $\pm$ 0.004 & 0.082 $\pm$ 0.001 & 0.203 $\pm$ 0.002 & 0.757 $\pm$ 0.001 & 0.821 $\pm$ 0.003 \\
Llama-3.2-3B & 30.51 $\pm$ 0.76 & 5.81 $\pm$ 0.10 & 0.251 $\pm$ 0.005 & 0.080 $\pm$ 0.003 & 0.203 $\pm$ 0.004 & 0.755 $\pm$ 0.001 & 0.779 $\pm$ 0.001 \\
Gemma-2-2B & 29.02 $\pm$ 0.62 & 4.78 $\pm$ 0.24 & 0.220 $\pm$ 0.004 & 0.064 $\pm$ 0.005 & 0.176 $\pm$ 0.004 & 0.750 $\pm$ 0.001 & 0.842 $\pm$ 0.002 \\
\bottomrule
\end{tabular}
\end{table}

\begin{figure}[tbp]
\centering
\includegraphics[width=5in]{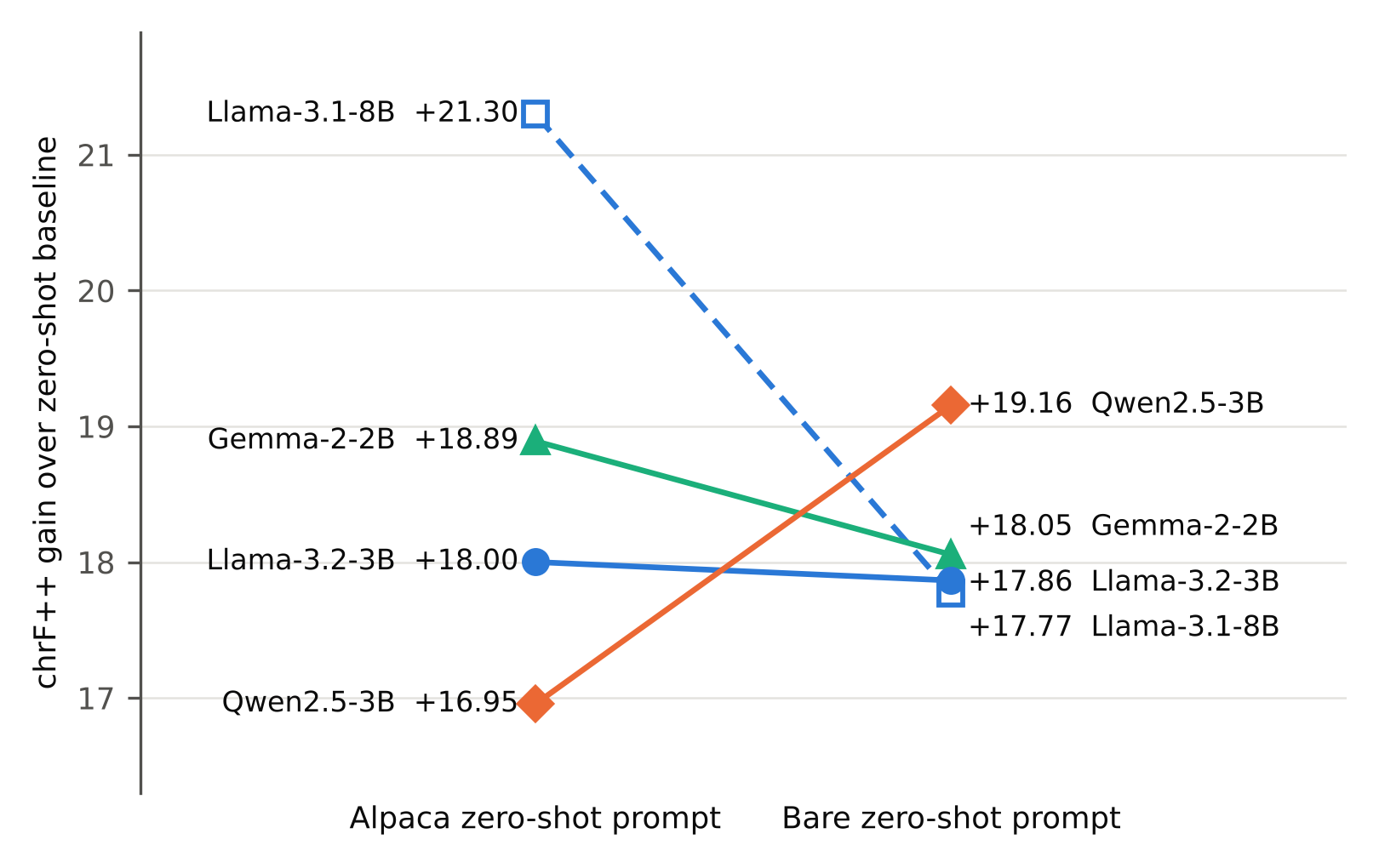}
\caption{Gain in chrF++ from zero-shot to fine-tuned, measured from the Alpaca-prompt and the bare-prompt zero-shot baselines. Lines connect the same model. Each model's fine-tuned score is the same in both columns; only the baseline changes.}\label{fig:gains}
\end{figure}

Gains over the zero-shot baselines depend on the baseline prompt (Table~\ref{tab:gains}, Figure~\ref{fig:gains}). Measured from the Alpaca-prompt baseline, chrF++ rises by 16.95 (Qwen2.5-3B) to 18.89 (Gemma-2-2B) points among the small models and by 21.30 for Llama-3.1-8B. Measured from the bare-prompt baseline, Qwen2.5-3B gains most (19.16) and Llama-3.1-8B least (17.77). The models with the largest and smallest gains thus swap places. Because each model's fine-tuned score is the same in both columns, these changes come entirely from the zero-shot runs. Under the bare prompt, three of the four gains lie within 0.3 points of one another, well inside seed variation, and BLEU gains give yet another order under each prompt. Likelihood gives a steadier picture: every model's fine-tuned BPB is 0.341–0.358 of its zero-shot value under either prompt, a 64–66\% reduction, and the BPB order of the four models is the same before and after fine-tuning.

\begin{table}[tbp]
\centering
\fontsize{8.5}{10}\selectfont
\setlength{\tabcolsep}{3pt}
\caption{Gains from zero-shot to fine-tuned: the three-seed fine-tuned mean minus the single zero-shot run under each prompt, and the ratio of fine-tuned to zero-shot BPB under each prompt. Computed from Tables~\ref{tab:zs}, \ref{tab:ft} and \ref{tab:8b} before rounding.}\label{tab:gains}
\begin{tabular}{@{}L{91.6pt}R{57.5pt}R{57.5pt}R{56.8pt}R{54.5pt}R{56.8pt}R{52.9pt}@{}}
\toprule
\textbf{Model} & \textbf{chrF++ gain, Alpaca} & \textbf{chrF++ gain, bare} & \textbf{BLEU gain, Alpaca} & \textbf{BLEU gain, bare} & \textbf{BPB ratio, Alpaca} & \textbf{BPB ratio, bare} \\
\midrule
Llama-3.1-8B & +21.30 & +17.77 & +4.99 & +6.00 & 0.358 & 0.349 \\
Gemma-2-2B & +18.89 & +18.05 & +4.37 & +4.66 & 0.349 & 0.341 \\
Llama-3.2-3B & +18.00 & +17.86 & +4.80 & +5.43 & 0.350 & 0.342 \\
Qwen2.5-3B & +16.95 & +19.16 & +5.22 & +5.42 & 0.346 & 0.342 \\
\bottomrule
\end{tabular}
\end{table}

\subsection{Training Stability and Seed Variation}\label{sec:res-seed}

Seed variation is of the same size as test-set variation. Seed SDs of chrF++ are 0.62–0.97 points, and the 95\% bootstrap interval of a single fine-tuned run is 1.7–2.0 points wide. Table~\ref{tab:diffs} shows what the two kinds of interval can resolve. Paired item-level comparisons resolve Gemma-2-2B's deficits in chrF++ and BLEU against both other small models and Llama-3.1-8B's small gains over Llama-3.2-3B, but not the difference between Llama-3.2-3B and Qwen2.5-3B (chrF++ $-$0.09 [$-$0.50, 0.31]). Seed-level intervals are wider: they resolve no chrF++ difference, and for BLEU only Gemma-2-2B's deficits and the 8B gain. Differences of about one chrF++ point between single runs are therefore not interpretable, and even three-seed means need a paired test. Training was stable apart from the one diverged Qwen2.5-3B replicate (Section~\ref{sec:ckpt}). After step 500, validation loss rose to at most 1.15 times its best earlier value for Gemma-2-2B and 1.115 times for Llama-3.1-8B; selection on the validation split keeps the earlier, better checkpoint.

\begin{table}[tbp]
\centering
\fontsize{8.5}{10}\selectfont
\setlength{\tabcolsep}{3pt}
\caption{Differences between models (first model minus second) with two kinds of 95\% interval. Paired item-level: the 1,000 test items resampled 2,000 times, scores recomputed for every seed and averaged over each model's seeds; it conditions on the trained seeds. Seed-level: Welch intervals over the three per-seed scores of each model; it conditions on the test set. Negative BPB differences mean higher likelihood for the first model. Intervals in bold exclude zero.}\label{tab:diffs}
\begin{tabular}{@{}L{146.7pt}L{45.5pt}R{60.5pt}R{93.5pt}R{93.5pt}@{}}
\toprule
\textbf{Comparison} & \textbf{Measure} & \textbf{Difference} & \textbf{Paired item-level 95\% CI} & \textbf{Seed-level 95\% CI} \\
\midrule
Llama-3.2-3B $-$ Qwen2.5-3B & BPB & $-$0.0423 & \textbf{[$-$0.0462, $-$0.0387]} & \textbf{[$-$0.0493, $-$0.0353]} \\
 & chrF++ & $-$0.09 & [$-$0.50, 0.31] & [$-$2.12, 1.93] \\
 & BLEU & $-$0.04 & [$-$0.56, 0.39] & [$-$0.62, 0.53] \\
Gemma-2-2B $-$ Qwen2.5-3B & BPB & +0.0205 & \textbf{[0.0168, 0.0244]} & \textbf{[0.0142, 0.0268]} \\
 & chrF++ & $-$1.59 & \textbf{[$-$2.01, $-$1.14]} & [$-$3.58, 0.40] \\
 & BLEU & $-$1.08 & \textbf{[$-$1.51, $-$0.68]} & \textbf{[$-$1.65, $-$0.50]} \\
Gemma-2-2B $-$ Llama-3.2-3B & BPB & +0.0628 & \textbf{[0.0580, 0.0684]} & \textbf{[0.0584, 0.0672]} \\
 & chrF++ & $-$1.50 & \textbf{[$-$1.92, $-$1.06]} & [$-$3.09, 0.10] \\
 & BLEU & $-$1.03 & \textbf{[$-$1.34, $-$0.75]} & \textbf{[$-$1.56, $-$0.51]} \\
Llama-3.1-8B $-$ Llama-3.2-3B & BPB & $-$0.0530 & \textbf{[$-$0.0559, $-$0.0501]} & \textbf{[$-$0.0571, $-$0.0490]} \\
 & chrF++ & +0.68 & \textbf{[0.26, 1.10]} & [$-$0.98, 2.33] \\
 & BLEU & +0.58 & \textbf{[0.17, 1.09]} & \textbf{[0.21, 0.95]} \\
\bottomrule
\end{tabular}
\end{table}

\subsection{Reference Likelihood against Generation Overlap}\label{sec:res-lik}

\begin{figure}[tbp]
\centering
\includegraphics[width=6.3in]{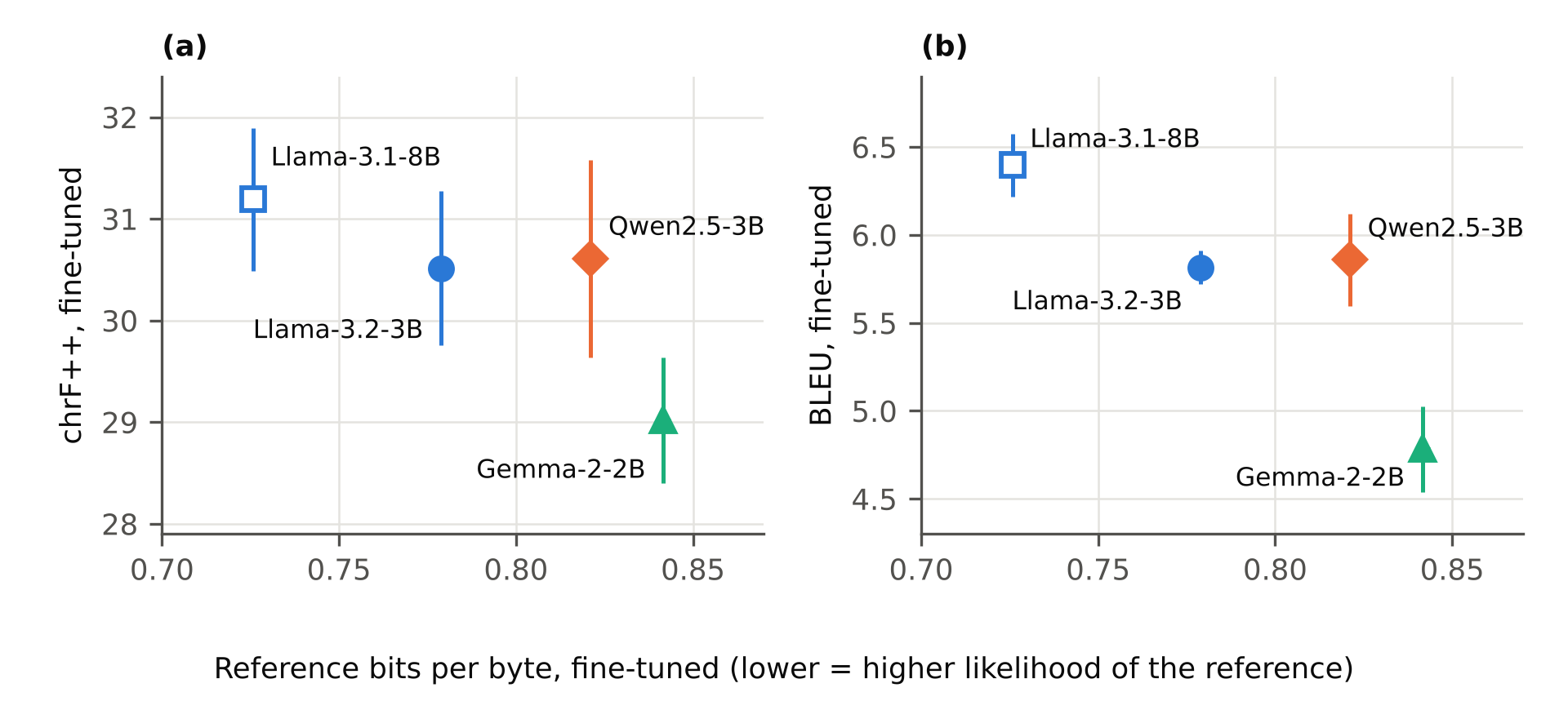}
\caption{Fine-tuned reference likelihood (BPB; lower means higher likelihood) against (a) chrF++ and (b) BLEU. Points are three-seed means and bars show $\pm$1 seed SD. Llama-3.2-3B and Qwen2.5-3B differ clearly in BPB but in neither generation metric.}\label{fig:bpb}
\end{figure}

Likelihood and generation overlap give different pictures of how the adapted models compare (Figure~\ref{fig:bpb}, Table~\ref{tab:diffs}). Every BPB difference between models is resolved by both kinds of interval, and the order holds in every seed. At the item level, the generation metrics separate every pair except one: Llama-3.2-3B and Qwen2.5-3B, the pair with the largest BPB gap among the small models (0.0423), cannot be separated on chrF++ or BLEU, nor on ROUGE or BERTScore. Gemma-2-2B's BPB deficit relative to Qwen2.5-3B is half as large (0.0205) but comes with the largest generation gap ($-$1.59 chrF++, $-$1.08 BLEU). The size of a likelihood gap therefore predicted neither the size nor the presence of a generation gap. By likelihood, Llama-3.2-3B is the best of the three small models; by generation overlap, it ties with Qwen2.5-3B. The two kinds of measure agree only on placing Gemma-2-2B last. Output length does not explain the pattern. Fine-tuned outputs were shorter than the references for every model (78–79\% of their length in characters on average), 3–8\% of them ran to the 256-token limit, and the two models that tie on generation also match on length (both 79\%). Because BPB is normalized by bytes, these differences are not an artefact of how many tokens each tokenizer produces, although they may still reflect tokenization together with the models' other differences.

\subsection{Within-Family Comparison with Llama-3.1-8B}\label{sec:res-8b}

Llama-3.1-8B tokenizes the references identically to Llama-3.2-3B, which removes segmentation as a source of difference between the two, although they differ in model generation, training data and distillation as well as size (Table~\ref{tab:models}). After fine-tuning (Table~\ref{tab:8b}), the 8B model has lower BPB in every seed, by 0.0530 [0.0501, 0.0559] in the paired comparison. Its gains in generation are small: 0.68 chrF++ and 0.58 BLEU points. Both are resolved by the paired item-level test (chrF++ [0.26, 1.10], BLEU [0.17, 1.09]), but only the BLEU gain is consistent across seeds: every 8B seed has higher BLEU than every 3B seed (6.20–6.54 against 5.72–5.91), whereas the 3B seed-43 run (chrF++ 31.38) scores above two of the three 8B runs. BERTScore differs by 0.005. The type A advantage holds in every 8B seed (chrF++ 32.47, 32.00 and 33.30 on type A against 29.15, 29.60 and 30.74 on type B). The audit shows no empty generations and no Devanagari output in the three 8B runs, with 0, 3 and 0 English-like outputs, 9, 13 and 9 input copies and 2, 2 and 4 repetition loops, similar to Llama-3.2-3B.

Without fine-tuning the likelihood gap is larger, but generation does not follow it. The 8B model's BPB is lower by 0.198 [0.192, 0.203] under the Alpaca prompt and by 0.194 [0.189, 0.199] under the bare prompt, while its chrF++ is lower under the Alpaca prompt (9.89 against 12.51) and higher under the bare prompt (13.42 against 12.65). Under the Alpaca prompt the 8B model returned 141 empty generations, 128 of them on type B items, against 4 for the 3B model; we did not investigate the cause. In sum, the larger model assigns higher probability to the references before and after fine-tuning, but after fine-tuning its outputs are only slightly closer to them, by 0.68 chrF++ and 0.58 BLEU points, and only the BLEU gain is consistent across seeds.

\begin{table}[tbp]
\centering
\fontsize{8.5}{10}\selectfont
\setlength{\tabcolsep}{3pt}
\caption{Llama-3.1-8B against Llama-3.2-3B. Fine-tuned rows: mean $\pm$ sample SD over seeds 42, 43 and 44 (8B per seed: chrF++ 30.79, 30.78 and 32.00; BLEU 6.20, 6.46 and 6.54; BPB 0.7263, 0.7274 and 0.7236). Zero-shot rows: single runs, with the BERTScore mean over non-empty outputs in parentheses. BPB is comparable across models within a condition, not across zero-shot prompts.}\label{tab:8b}
\begin{tabular}{@{}L{87.6pt}L{67.7pt}R{56.9pt}R{52.0pt}R{69.7pt}R{61.2pt}R{32.7pt}@{}}
\toprule
\textbf{Condition} & \textbf{Model} & \textbf{chrF++} & \textbf{BLEU} & \textbf{BERTScore F1} & \textbf{BPB} & \textbf{Empty} \\
\midrule
Fine-tuned & Llama-3.1-8B & 31.19 $\pm$ 0.70 & 6.40 $\pm$ 0.18 & 0.760 $\pm$ 0.0004 & 0.726 $\pm$ 0.002 & 0 \\
Fine-tuned & Llama-3.2-3B & 30.51 $\pm$ 0.76 & 5.81 $\pm$ 0.10 & 0.755 $\pm$ 0.001 & 0.779 $\pm$ 0.001 & 0 \\
Zero-shot, Alpaca & Llama-3.1-8B & 9.89 & 1.41 & 0.500 (0.583) & 2.027 & 141 \\
Zero-shot, Alpaca & Llama-3.2-3B & 12.51 & 1.01 & 0.587 & 2.225 & 4 \\
Zero-shot, bare & Llama-3.1-8B & 13.42 & 0.40 & 0.533 (0.541) & 2.082 & 14 \\
Zero-shot, bare & Llama-3.2-3B & 12.65 & 0.38 & 0.536 & 2.276 & 9 \\
\bottomrule
\end{tabular}
\end{table}

\section{Discussion}\label{sec:discussion}

\subsection{What Likelihood and Generation Overlap Each Show}

Across the four adapted models, BPB separated every pair. The overlap metrics separated every pair except Llama-3.2-3B and Qwen2.5-3B, which had the largest likelihood gap among the small models. The two kinds of measure answer different questions. BPB measures how probable a model finds one human-written response under teacher forcing; chrF++ and BLEU measure how close the model's own greedy output is to that response.

Several mechanisms could produce the gap, and our design does not test them. Teacher-forced scoring conditions each token on the reference prefix, whereas free-running greedy decoding conditions on the model's own earlier choices, so a model can rank the reference well without producing it. With one reference per item, many adequate responses score poorly on overlap, which adds noise that likelihood does not have. Greedy decoding with a 256-token limit and a repetition penalty changes the generated text in ways that likelihood scoring does not see, and the 9–12\% of references that are longer than the limit cannot be matched in full by any model; output length itself, however, was similar across models (Section~\ref{sec:res-lik}). Distillation could also play a part: Llama-3.2-3B was pretrained with token-level targets from larger Llama models \citep{meta2024b}. Gemma-2-2B was also distilled \citep{gemma2024} yet has the highest BPB, so distillation alone does not explain the pattern.

Meister and Cotterell \citep{meister2021} argued that perplexity does not characterize the text a model generates; with byte normalization, the dissociation holds even though likelihood is compared across different tokenizers. The practical consequence is that neither kind of measure can stand in for the other: a study that selected among our models by likelihood would have preferred Llama-3.2-3B over Qwen2.5-3B, and one that selected by chrF++ would have found no difference. The evidence comes from four models and one test set.

\subsection{What Tokenizer Fertility Can and Cannot Indicate}\label{sec:disc-fert}

Fertility ranked the tokenizers consistently, on the benchmark references and under three romanization conventions, but it did not identify the best model: the tokenizer that split words least belonged to the model that scored lowest. Three points limit what this shows. The range among the trained models was narrow (2.72–3.00 tokens per word), far from the many-fold differences between languages \citep{petrov2023} or between native and romanized scripts \citep{husain2024} for which fertility is known to matter. The models differ in much more than the tokenizer (Table~\ref{tab:models}). And fertility itself depended on the romanization convention about as much as on the tokenizer (Section~\ref{sec:res-fert}).

The result agrees with controlled pretraining studies in which fertility or token count did not reliably predict downstream performance \citep{ali2024,schmidt2024}, and with transfer experiments in which extending the vocabulary to shorten target-language text was not needed \citep{zhao2024}. It does not contradict the finding that compression tracks performance when everything except the tokenizer is held fixed \citep{goldman2024}. Our measurements also do not support explaining Llama-family results by over-fragmentation \citep{rimal2026}: on the same text, the Llama 3 tokenizer split words into fewer pieces than the Qwen and Mistral tokenizers.

One mechanism deserves attention in future work. A large vocabulary lowers fertility but costs embedding parameters. Gemma-2-2B devotes 22.6\% of its parameters to the embeddings of its 256k vocabulary and has the fewest non-embedding parameters of the three models (2.02B against 2.77–2.82B), so in our models lower fertility and lower non-embedding capacity come together, and the data cannot separate them.

\subsection{Interpreting Gemma-2-2B}

Gemma-2-2B is not an adaptation failure. Its chrF++ gain under the Alpaca prompt (+18.89) is the largest of the three small models, and its BPB falls by the same proportion as the others' (to 0.349 of its zero-shot value). It starts lowest and ends lowest. Several differences could contribute, and our design cannot separate them: the fewest non-embedding parameters and pretraining tokens of the three models (Table~\ref{tab:models}); pretraining data that are primarily English \citep{gemma2024}; attention and final-logit soft-capping; Unsloth compilation disabled; a recipe tuned for none of the models; and a late rise in validation loss to 1.14–1.15 times its best earlier value. Its zero-shot behaviour, with the most English-like output across the two prompts and the lowest zero-shot chrF++ of the three small models, fits less pretraining exposure to Romanized Nepali, but exposure was not measured.

\subsection{Zero-Shot Baselines and Gains}

On this task, zero-shot baselines measure language choice and prompt-following as much as ability. Depending on the model and prompt, the base models answered in English, copied the prompt, looped or returned nothing. Gains over these baselines therefore mix a switch of output language with task learning, and the order of models by gain changed with the prompt and with the metric (Table~\ref{tab:gains}). Rimal and Rimal \citep{rimal2026} read such gains as evidence that models with weaker zero-shot performance have more representational plasticity. Our data do not support ranking models by gains when the base models mostly answer in another language: the model with the largest chrF++ gain under one prompt had the smallest under the other. The proportional fall in BPB was nearly constant across the four models and both prompts (64–66\%), which suggests that likelihood is a steadier summary of how much a model changes with adaptation, within the limits discussed above. The concentration of zero-shot failures on type B prompts is consistent with base models having difficulty reading Romanized Nepali instructions.

\subsection{The Within-Family Comparison}

With the tokenizer held fixed, the larger Llama model had clearly higher reference likelihood before and after fine-tuning, but after fine-tuning its outputs were only slightly closer to the references: by 0.68 chrF++ and 0.58 BLEU points, gains that hold across test items but, for chrF++, not across seeds. The comparison thus repeats, within one family and without any difference in segmentation, the pattern seen across families: a likelihood advantage that carries over to generation only in part. It cannot isolate the effect of parameter count, because Llama-3.1-8B and Llama-3.2-3B differ in model generation, data and distillation as well as size; Llama-3.2-3B was trained in part on the larger models' output distributions \citep{meta2024b}.

\subsection{Implications}

Four practical points follow for adapting models to romanized varieties such as Romanized Nepali. First, tokenizer fertility should not be used on its own to choose among pretrained models; within the narrow range of modern tokenizers, other differences dominated. Second, the romanization convention is a design variable. In our measurements, lowercasing changed fertility by 13–14\% and dropping the final inherent vowel by a further 3–4\%, and the convention decides which distinctions of the script a model can recover, so benchmarks should report it. Third, adapted models should be evaluated with a byte-normalized likelihood together with generation metrics and an output audit, because the two kinds of measure separate models differently. Fourth, comparisons need several seeds and paired tests: chrF++ differences of 1.5–1.6 points were resolved by paired item-level tests but not across three seeds, and neither kind of interval alone captures both sources of variation.

\section{Limitations}\label{sec:limitations}

\textbf{Data.} All results come from one instruction dataset of 10,500 examples, derived from a Nepali translation of the Stanford Alpaca data whose dataset card documents neither its provenance nor a licence. Romanized text was produced by deterministic, lowercased ITRANS transliteration. This scheme keeps word-final inherent vowels, merges 11 groups of letters and gives every word one spelling, so the benchmark does not test robustness to the spelling variation of naturally typed Romanized Nepali, to mixed scripts or to code-switching. Type A instructions were machine-translated back into English from a Nepali translation of English originals, which sometimes changes their meaning, and the two types differ in how the instruction was produced as well as in its language. Some training examples (1.4–2.0\%) were truncated at 512 tokens, and a few outputs recur across splits.

\textbf{Models and recipe.} Three base-model families were trained with one shared recipe that was not tuned per model. The models differ in tokenizer, architecture, non-embedding parameter count, pretraining data and distillation at once, so no difference between them can be attributed to any single factor, including the tokenizer. Unsloth left different modules unquantized in different models.

\textbf{Evaluation.} All measures are automatic and single-reference. BLEU, chrF++ and ROUGE reward overlap with one reference; BERTScore relies on a multilingual encoder that has not been validated for Romanized Nepali; and bits per byte measures how probable the reference is, not how good a generated response is. The output audit uses surface heuristics, and its English-stopword list was screened against the test references. Zero-shot results depend on two prompt formats and our decoding settings.

\textbf{Statistics.} Three seeds give only a rough estimate of training variability. The paired item-level intervals condition on the trained seeds and the seed-level intervals on the test set; neither captures both sources of variation at once. BERTScore could not be recomputed from the saved outputs, so its values are those of the evaluation scripts. The divergence criterion was set post hoc, and the selected checkpoint steps were not retained.

\textbf{Fertility analysis.} The public-sample analysis used tokenizer files rebuilt from public sources, which reproduce the Hugging Face token counts exactly on the benchmark references, and a simple heuristic for dropping the final inherent vowel. Its sentences have shorter words than the benchmark responses.

\textbf{Within-family comparison.} Llama-3.1-8B and Llama-3.2-3B differ in model generation, training data and distillation as well as size. The comparison does not measure an effect of parameter count.

\textbf{Software.} Package versions differed across runs. One re-evaluation in a second environment reproduced its results exactly, but it covers one adapter.

Independent human assessment and evaluation on naturally written Romanized Nepali prompts are the main directions for future work.

\section{Conclusion}\label{sec:conclusion}

We asked how small open language models adapt to Romanized Nepali and whether common shortcuts identify the best one. All three 2–3B families adapted substantially under a shared QLoRA recipe, and their results varied across seeds by about as much as across test samples (RQ1). The shortcuts did not hold. The tokenizer that fragmented Romanized Nepali least belonged to the lowest-scoring model (RQ2). The model that assigned the highest likelihood to the references did not generate text closer to them than its nearest competitor, and within the Llama family, where tokenization was identical, the larger model's likelihood advantage carried over to generation only in part (RQ3). Without adaptation the base models mostly answered in English, and which model gained most from fine-tuning depended on the zero-shot prompt (RQ4). For low-resource and romanized varieties, where a single proxy is tempting, these results argue for evaluating adapted models with several kinds of measure and explicit uncertainty. These conclusions hold for the transliterated benchmark on which they were obtained.

\section*{Ethics Statement}\label{sec:ethics}

We used a publicly available instruction dataset whose card states no licence. It is a Nepali translation of the Stanford Alpaca data (Section~\ref{sec:source}), whose CC~BY-NC~4.0 licence restricts it to non-commercial use. We collected no personal data and recruited no participants. The sentences used in the fertility analysis come from the Mozilla Common Voice text corpus, released under CC0, and a Nepali translation of the Universal Declaration of Human Rights distributed with NLTK. Models were used under their licences: the Llama 3.2 and Llama 3.1 Community Licenses, the Gemma Terms of Use and the Qwen Research License, which restricts Qwen2.5-3B to research use. Fine-tuned models can produce incorrect or biased text, and even the fine-tuned models occasionally copied prompts, looped or switched to English, so they should not be used in high-stakes settings without human review. The transliteration scheme determines which spellings the models learn and does not reflect how all writers of Romanized Nepali spell.

\section*{Data and Code Availability}\label{sec:availability}

The training and evaluation scripts, the code and results of the analyses in this paper, and the per-item generations, reference NLL values and token and byte counts of the 19 evaluation runs whose outputs were retained are available at \url{https://github.com/pantpujan017/romanized-nepali-slm-adaptation}. Because the source dataset states no licence, we release model outputs rather than its text.

\section*{Acknowledgments}\label{sec:ack}

During the preparation of this study, the authors used Claude Opus 5.5 (Anthropic) to draft and refine the manuscript text, to check references, to write the code for the re-analysis of the saved outputs (Tables~\ref{tab:gains} and \ref{tab:diffs} and Appendix~\ref{app:seeds}) and for the tokenizer and romanization analysis (Section~\ref{sec:res-fert} and Appendix~\ref{app:roman}), and to prepare the plotting code for Figures~\ref{fig:gains} and \ref{fig:bpb}. Claude (Anthropic) was also used to write the code for the experiments, including the training and evaluation scripts. The authors have reviewed and edited the output and take full responsibility for the content of this publication.

\clearpage
\appendix

\section{Run Provenance}\label{app:prov}

Hashes are SHA-256 prefixes; full hashes are in the code repository. “Not recorded” marks information for which no run artifact was retained.

\begin{table}[H]
\centering
\fontsize{8.5}{10}\selectfont
\setlength{\tabcolsep}{3pt}
\caption{Scripts, environments and records behind the reported numbers.}\label{tab:prov}
\begin{tabular}{@{}L{123.2pt}L{334.5pt}@{}}
\toprule
\textbf{Item} & \textbf{Detail} \\
\midrule
Dataset splits & train 9ed86da7\ldots{}, validation 3f7f6854\ldots{}, test d656353e\ldots{}; hash-identical copies of the dataset directory were uploaded to Kaggle more than once and read by different sessions and accounts (the zero-shot launch records and the 8B seed-44 run use a second copy); all 19 retained runs have the same 1,000 test references. \\
Tokenizer revisions & Fertility and length analyses (Transformers 5.16.1, tokenizers 0.23.1): Llama-3.2-3B 13afe512\ldots{}, Qwen2.5-3B 3aab1f19\ldots{}, gemma-2-2b c5ebcd40\ldots{}, Llama-3.1-8B d04e592b\ldots{}, Mistral-7B-v0.1 27d67f1b\ldots{}, Qwen3-8B b968826d\ldots{}, Qwen3-8B-Base 49e3418f\ldots{}; fertility script a0c856c5\ldots{}, length-audit script 3938c8a4\ldots{}. \\
Model weights & Loaded through Unsloth, which substitutes its pre-quantized 4-bit checkpoints (Section~\ref{sec:recipe}); their revisions were not recorded. \\
Training, primary models & train.py v2 (579d36c7\ldots{}, canonical) with per-run patched copies; Qwen2.5-3B seeds 43 and 44 were trained in parallel, one per GPU, with Unsloth 2026.10.11, from a copy that differs only in its CONFIG values and one comment line (d9d03919\ldots{}); the excluded second seed-42 replicate used train.py v1, which has the same training settings, with Unsloth 2026.10.11 and Transformers 5.16.1; other Unsloth versions, bitsandbytes and TRL versions not recorded. \\
Training, Llama-3.1-8B & canonical train.py with only the model, data and output paths changed; patched copies b6949e2f\ldots{} (seed 42), d6212c1f\ldots{} (seed 43) and 662c1093\ldots{} (seed 44); Unsloth 2026.11.3; seed 44 trained in a separate Kaggle account from a hash-identical copy of the data. \\
Fine-tuned evaluation, primary models & eval.py v2.2 (metric files record the script hash as unknown); Llama-3.2-3B evaluations used Unsloth 2026.10.11 (seed 42) and 2026.11.1 (seeds 43 and 44) with Transformers 5.17.0, PEFT 0.21.2 and torch 2.11.0+cu130; Qwen2.5-3B seed 42 was evaluated with Unsloth 2026.10.11 and Transformers 5.16.1, and seeds 43 and 44 with a copy of eval.py v2.2 whose hash was recorded (a231a1ae\ldots{}); other Qwen2.5-3B and the Gemma-2-2B evaluation versions not recorded. \\
Zero-shot evaluation, primary models & one script (33e10eac\ldots{}) with per-run patched copies. Qwen2.5-3B Alpaca: patched-copy hash not recorded, Unsloth 2026.11.2. Qwen2.5-3B bare: 618c89fb\ldots{}, started 2026-11-06 02:39:01 UTC, Unsloth 2026.11.2. Llama-3.2-3B Alpaca: 11191d0c\ldots{}, started 2026-11-07 01:25:20 UTC, Unsloth 2026.11.3. Llama-3.2-3B bare: c8a28578\ldots{}, Unsloth 2026.11.2; launch records give 2026-11-06 08:02:48 UTC (recovery record) and 12:51:30 UTC (launch file). Both Gemma-2-2B runs: Unsloth 2026.11.3. \\
Metric recovery & recover\_metrics.py (ba4a20c5\ldots{}), used for zero-shot runs with empty generations, which keeps empty hypotheses at BERTScore 0; validated on the Qwen2.5-3B Alpaca run (Section~\ref{sec:metrics}). \\
Llama-3.1-8B evaluation & eval.py v2.4 (51e1bd41\ldots{}); patched copies f34da070\ldots{} (seed 42), 12c1f7c3\ldots{} (seed 43) and c288f295\ldots{} (seed 44); Unsloth 2026.11.3, Transformers 5.17.0, PEFT 0.21.2 and torch 2.11.0+cu130; launch times not recorded. \\
Llama-3.1-8B zero-shot & Alpaca: patched copy a587b947\ldots{}, started 2026-11-08 04:50:24 UTC; bare: generation script 7f36ac2f\ldots{}; metrics rebuilt with recover\_metrics.py. The bare-prompt outputs were not retained. \\
Output audit & output\_audit.py (69b88899\ldots{}), applied to all 20 evaluation runs and the reference responses. \\
Re-evaluation check & Qwen2.5-3B seed 43 re-evaluated with Unsloth 2026.11.2, Transformers 5.17.0, PEFT 0.21.2 and PyTorch for CUDA 13.0: identical generations, per-item NLL values and metrics. \\
Metric signatures & BLEU \texttt{nrefs:1|case:mixed|eff:no|tok:13a|smooth:exp|version:2.6.0}; chrF++ \texttt{nrefs:1|case:mixed|eff:yes|nc:6|nw:2|space:no|version:2.6.0}; evaluation-script intervals used 200 resamples with Python random seed 0, re-initialized for each subset. \\
Re-analysis & run\_level\_analysis.py in the code repository (sacreBLEU 2.6.0, rouge-score 0.1.2), applied to the per-item outputs of 19 runs; it reproduces every reported chrF++, BLEU, ROUGE and BPB value. \\
\bottomrule
\end{tabular}
\end{table}

\begin{table}[H]
\centering
\fontsize{8.5}{10}\selectfont
\setlength{\tabcolsep}{3pt}
\caption{Training runs and their use. Checkpoints were selected on validation loss (Section~\ref{sec:ckpt}); the retained records give the selected step only where noted.}\label{tab:runs}
\begin{tabular}{@{}L{72.2pt}R{28.6pt}L{95.1pt}L{63.7pt}L{180.3pt}@{}}
\toprule
\textbf{Model} & \textbf{Seed} & \textbf{Status} & \textbf{Unsloth (training)} & \textbf{Notes} \\
\midrule
Qwen2.5-3B & 42 & used (first run) & not recorded &  \\
Qwen2.5-3B & 42 & excluded (diverged) & 2026.10.11 & second replicate, train.py v1 \\
Qwen2.5-3B & 43 & used & 2026.10.11 & selected step 2200 \\
Qwen2.5-3B & 44 & used & 2026.10.11 & selected step 2200 \\
Llama-3.2-3B & 42 & used & not recorded &  \\
Llama-3.2-3B & 43 & used & not recorded & one non-finite gradient norm (step 2150) \\
Llama-3.2-3B & 44 & used & not recorded &  \\
Gemma-2-2B & 42 & used & not recorded & compilation disabled \\
Gemma-2-2B & 43 & used & not recorded & compilation disabled \\
Gemma-2-2B & 44 & used & not recorded & compilation disabled \\
Llama-3.1-8B & 42 & used & 2026.11.3 &  \\
Llama-3.1-8B & 43 & used & 2026.11.3 &  \\
Llama-3.1-8B & 44 & used & 2026.11.3 & separate Kaggle account \\
\bottomrule
\end{tabular}
\end{table}

\section{Romanization and Fertility Analysis}\label{app:roman}

\textbf{Corpus.} The public sample combines 1,146 sentences from the Mozilla Common Voice Nepali text corpus (CC0) \citep{ardila2020} and 87 lines of the Nepali Universal Declaration of Human Rights from NLTK's udhr2 corpus \citep{bird2009}, after Unicode NFC normalization and removal of duplicates and lines without Devanagari. It has 8,840 whitespace-delimited words; the 8,102 that remain after removing standalone punctuation are used for the lexical statistics in Section~\ref{sec:roman}.

\textbf{Conventions.} Devanagari was converted to ITRANS with indic\_transliteration 2.3.82 \citep{indictrans} (R1), then lowercased (R2). R3 removes the final \emph{a} from R2 words whose Devanagari form ends in a consonant letter without a virama and has at least two aksharas; words ending in a candrabindu, anusvara or visarga are unchanged. Trailing punctuation is split off before transliteration, so the danda becomes “|”. Under R2, the 11 groups in Table~\ref{tab:merged} merge; lowercasing also merges the anusvara with a virama-marked \dev{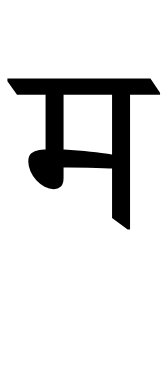} (both \emph{m}) and the visarga with a virama-marked \dev{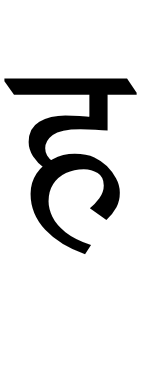} (both \emph{h}), and it merges the nukta letters \dev{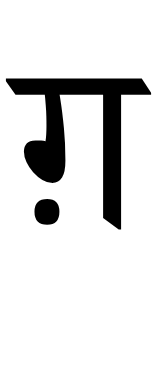}, \dev{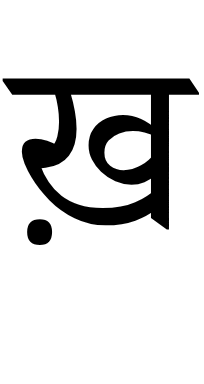}, \dev{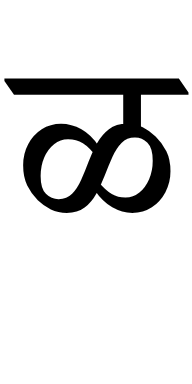} and \dev{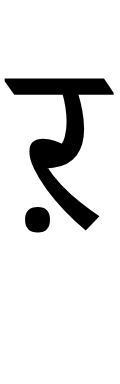} with their plain forms, although these are rare in Nepali.

\begin{table}[H]
\centering
\fontsize{8.5}{10}\selectfont
\setlength{\tabcolsep}{3pt}
\caption{Letter groups merged by lowercasing ITRANS. Vowel groups include the corresponding vowel signs.}\label{tab:merged}
\begin{tabular}{@{}L{133.0pt}L{185.7pt}L{133.0pt}@{}}
\toprule
\textbf{Devanagari} & \textbf{ITRANS (case-preserving)} & \textbf{Lowercased} \\
\midrule
\dev{deva/d12.png} / \dev{deva/d13.png} & \texttt{ch} / \texttt{Ch} & \texttt{ch} \\
\dev{deva/d21.png} / \dev{deva/d15.png} & \texttt{t} / \texttt{T} & \texttt{t} \\
\dev{deva/d24.png} / \dev{deva/d17.png} & \texttt{th} / \texttt{Th} & \texttt{th} \\
\dev{deva/d25.png} / \dev{deva/d18.png} & \texttt{d} / \texttt{D} & \texttt{d} \\
\dev{deva/d26.png} / \dev{deva/d19.png} & \texttt{dh} / \texttt{Dh} & \texttt{dh} \\
\dev{deva/d27.png} / \dev{deva/d20.png} & \texttt{n} / \texttt{N} & \texttt{n} \\
\dev{deva/d37.png} / \dev{deva/d38.png} & \texttt{sh} / \texttt{Sh} & \texttt{sh} \\
\dev{deva/d14.png} / \dev{deva/d11.png} & \texttt{\textasciitilde{}n} / \texttt{\textasciitilde{}N} & \texttt{\textasciitilde{}n} \\
\dev{deva/d01.png} / \dev{deva/d02.png} & \texttt{a} / \texttt{A} & \texttt{a} \\
\dev{deva/d03.png} / \dev{deva/d04.png} & \texttt{i} / \texttt{I} & \texttt{i} \\
\dev{deva/d05.png} / \dev{deva/d06.png} & \texttt{u} / \texttt{U} & \texttt{u} \\
\bottomrule
\end{tabular}
\end{table}

\textbf{Tokenizers.} Hugging Face model files could not be retrieved in the analysis environment, so tokenizers were rebuilt from public files with each model's published pre-tokenization: the Llama 3 BPE ranks from Meta's llama-models repository with the Llama 3 split pattern, the Qwen BPE ranks (151,643 tokens) distributed with the dashscope 2.0.4 package with Qwen's pattern (single digits), the Gemma SentencePiece model from the gemma\_pytorch repository, and the Mistral v0.1 SentencePiece model from mistral-common 1.13.1. Word-level counts encode each word with a leading space, or with SentencePiece's own prefix for Mistral. As a check, these tokenizers reproduce the Hugging Face token counts of the Llama-3.2-3B, Qwen2.5-3B, gemma-2-2b and Mistral-7B-v0.1 tokenizers exactly on all 1,000 benchmark references. Fertility is the mean over sentences of tokens per word, with 1,000 bootstrap resamples over sentences (seed 0); the order of tokenizers was checked over 2,000 paired resamples (seed 1). The script and its full output, including intervals for all conventions and type-level fertility, are in the code repository.

\section{Per-Seed Results}\label{app:seeds}

Table~\ref{tab:seeds} gives every fine-tuned run, recomputed from the saved outputs. The values reproduce those of the evaluation scripts, and the means and SDs of Tables~\ref{tab:ft} and \ref{tab:8b} follow from them.

\begin{table}[H]
\centering
\fontsize{8.5}{10}\selectfont
\setlength{\tabcolsep}{3pt}
\caption{Per-seed results of the fine-tuned runs on the 1,000 test items. Length is total generated characters divided by total reference characters.}\label{tab:seeds}
\begin{tabular}{@{}L{76.8pt}R{29.9pt}R{42.9pt}R{37.9pt}R{57.6pt}R{39.7pt}R{44.9pt}R{44.9pt}R{41.2pt}@{}}
\toprule
\textbf{Model} & \textbf{Seed} & \textbf{chrF++} & \textbf{BLEU} & \textbf{ROUGE-L} & \textbf{BPB} & \textbf{chrF++ type A} & \textbf{chrF++ type B} & \textbf{Length} \\
\midrule
Qwen2.5-3B & 42 & 29.53 & 5.61 & 0.202 & 0.8208 & 30.70 & 28.39 & 0.75 \\
Qwen2.5-3B & 43 & 31.42 & 5.84 & 0.201 & 0.8243 & 32.68 & 30.21 & 0.84 \\
Qwen2.5-3B & 44 & 30.87 & 6.13 & 0.205 & 0.8182 & 32.16 & 29.61 & 0.79 \\
Llama-3.2-3B & 42 & 30.19 & 5.82 & 0.201 & 0.7795 & 31.67 & 28.75 & 0.79 \\
Llama-3.2-3B & 43 & 31.38 & 5.91 & 0.200 & 0.7779 & 32.64 & 30.15 & 0.83 \\
Llama-3.2-3B & 44 & 29.97 & 5.72 & 0.207 & 0.7790 & 31.45 & 28.53 & 0.76 \\
Gemma-2-2B & 42 & 28.68 & 4.51 & 0.175 & 0.8432 & 29.86 & 27.55 & 0.78 \\
Gemma-2-2B & 43 & 28.64 & 4.99 & 0.180 & 0.8424 & 29.89 & 27.43 & 0.76 \\
Gemma-2-2B & 44 & 29.73 & 4.84 & 0.172 & 0.8393 & 30.99 & 28.51 & 0.81 \\
Llama-3.1-8B & 42 & 30.79 & 6.20 & 0.211 & 0.7263 & 32.47 & 29.15 & 0.76 \\
Llama-3.1-8B & 43 & 30.78 & 6.46 & 0.208 & 0.7274 & 32.00 & 29.60 & 0.76 \\
Llama-3.1-8B & 44 & 32.00 & 6.54 & 0.212 & 0.7236 & 33.30 & 30.74 & 0.82 \\
\bottomrule
\end{tabular}
\end{table}

\end{document}